\documentclass[nohyperref]{article}

\usepackage{microtype}
\usepackage{graphicx}
\usepackage{subfigure}
\usepackage{booktabs} 

\usepackage{hyperref}

\usepackage[accepted]{icml2022}

\usepackage{amssymb}
\usepackage{mathtools}
\usepackage{amsthm}
\usepackage{bm}

\DeclareMathOperator*{\argmax}{arg\,max}

\usepackage[capitalize,noabbrev]{cleveref}

\theoremstyle{plain}

\theoremstyle{definition}

\theoremstyle{remark}

\usepackage[textsize=tiny]{todonotes}

\usepackage{amsmath}
\icmltitlerunning{Reconstructing Nonlinear Dynamical Systems from Multi-Modal Time Series}

\begin{document}

\twocolumn[
\icmltitle{Reconstructing Nonlinear Dynamical Systems from Multi-Modal Time Series}



\icmlsetsymbol{equal}{*}

\begin{icmlauthorlist}
\icmlauthor{Daniel Kramer,}{1,equal}
\icmlauthor{Philine Lou Bommer,}{2,equal}
\icmlauthor{Carlo Tombolini,}{1}
\icmlauthor{Georgia Koppe}{1,4}
\icmlauthor{and Daniel Durstewitz}{1,3}

\end{icmlauthorlist}

\icmlaffiliation{1}{Dept. of Theoretical Neuroscience, Central Institute of Mental Health, Heidelberg University, Germany.}
\icmlaffiliation{2}{Dept. of Machine Learning, Technical University Berlin, Berlin, Germany.}
\icmlaffiliation{3}{Faculty of Physics and Astronomy, Heidelberg University, Germany.}
\icmlaffiliation{4}{Clinic for Psychiatry and Psychotherapy, Central Institute of Mental Health, Mannheim, Germany.}

\icmlcorrespondingauthor{Daniel Durstewitz}{daniel.durstewitz@zi-mannheim.de}
\icmlcorrespondingauthor{Philine Lou Bommer}{philine.l.bommer@tu-berlin.de}

\icmlkeywords{Machine Learning, ICML}

\vskip 0.3in
]


\printAffiliationsAndNotice{\icmlEqualContribution} 

\begin{abstract}
Empirically observed time series in physics, biology, or medicine, are commonly generated by some underlying dynamical system (DS) which is the target of scientific interest. There is an increasing interest to harvest machine learning methods to reconstruct this latent DS in a data-driven, unsupervised way. In many areas of science it is common to sample time series observations from many data modalities simultaneously, e.g. electrophysiological and behavioral time series in a typical neuroscience experiment. However, current machine learning tools for reconstructing DSs usually focus on just one data modality. Here we propose a general framework for multi-modal data integration for the purpose of nonlinear DS reconstruction and the analysis of cross-modal relations. This framework is based on dynamically interpretable recurrent neural networks as general approximators of nonlinear DSs, coupled to sets of modality-specific decoder models from the class of generalized linear models. Both an expectation-maximization and a variational inference algorithm for model training are advanced and compared. 
We show on nonlinear DS benchmarks that our algorithms can efficiently compensate for too noisy or missing information in one data channel by exploiting other channels, and demonstrate on experimental neuroscience data how the algorithm learns to link different data domains to the underlying dynamics.\\

\end{abstract}

\section{Introduction}
\label{intro}

Many natural phenomena, from physics to psychology, as well as many engineered systems, can genuinely be described as (usually nonlinear) dynamical systems (DSs), whose temporal evolution is specified by a set of differential or time-recursive equations. While traditionally these systems are derived by scientific insight, in recent years there has been growing interest to infer the governing equations directly from time series observations, in a purely data-driven way, using machine learning tools, such as polynomial regression \cite{Brunton16,Champion19}, Gaussian processes \cite{Duncker19}, or recurrent neural networks (RNNs) \cite{Lu17,Durstewitz17,Koppe19,hernandez2018, Vlachas2018, Pathak2018}. Based on Cybenko’s universal approximation theorem \cite{Cybenko89}, it has been shown that RNNs with sigmoid \cite{Funahashi93,Kimura98, Hanson2020} or Rectified Linear Unit (ReLU) \cite{LuWang17,Lin18} activation functions are theoretically powerful enough to approximate any DS, i.e. its vector field, to arbitrary precision in its own set of dynamical equations \cite{Funahashi93,Trischler16}. 
The objective of reconstructing, or approximating, the underlying DS itself, is more challenging compared to the more common goal of training a system to produce good ahead-predictions of temporal sequences \cite{Koppe19}. This is because in DS reconstruction we require 
the trained model to also reproduce \textit{invariant} properties of the underlying system in the limit $t \rightarrow \infty$, including the geometrical structure of its limit sets (attractors) or its temporal structure as assessed by the power spectrum.

Many natural and engineered DSs can be observed through many different measurement channels that produce time series: 
In Smartphone apps tracking psychiatric risk and behavior, for instance, one may want to combine categorical or ordinal information from ecological momentary assessments (EMA) with different continuous sensor readings, typing dynamics, proxies for social interactions, and GPS tracking \cite{Radu16,Koppe18}. In typical experiments in neuroscience one observes at the same time both continuous, often high-dimensional, measurements from the brain by means of electrophysiological or neuroimaging tools, and a subject's often categorical behavioral responses across many trials. Not only is it often desirable to directly relate these different data streams within a common latent model, e.g. to gain insight into how neural activity produces behavior, or to predict behavioral choices from accelerometer readings, but the different data streams may also convey complementary information about the underlying DS that will supplement each other and help in identifying the system. Yet, different types of time series data may require very different distributional assumptions, especially when dealing with both continuous or ordinal data and categorical, event-type information.

\begin{figure*}[tbp]
    \centering
    \begin{center}
    \includegraphics[width=0.75\textwidth]{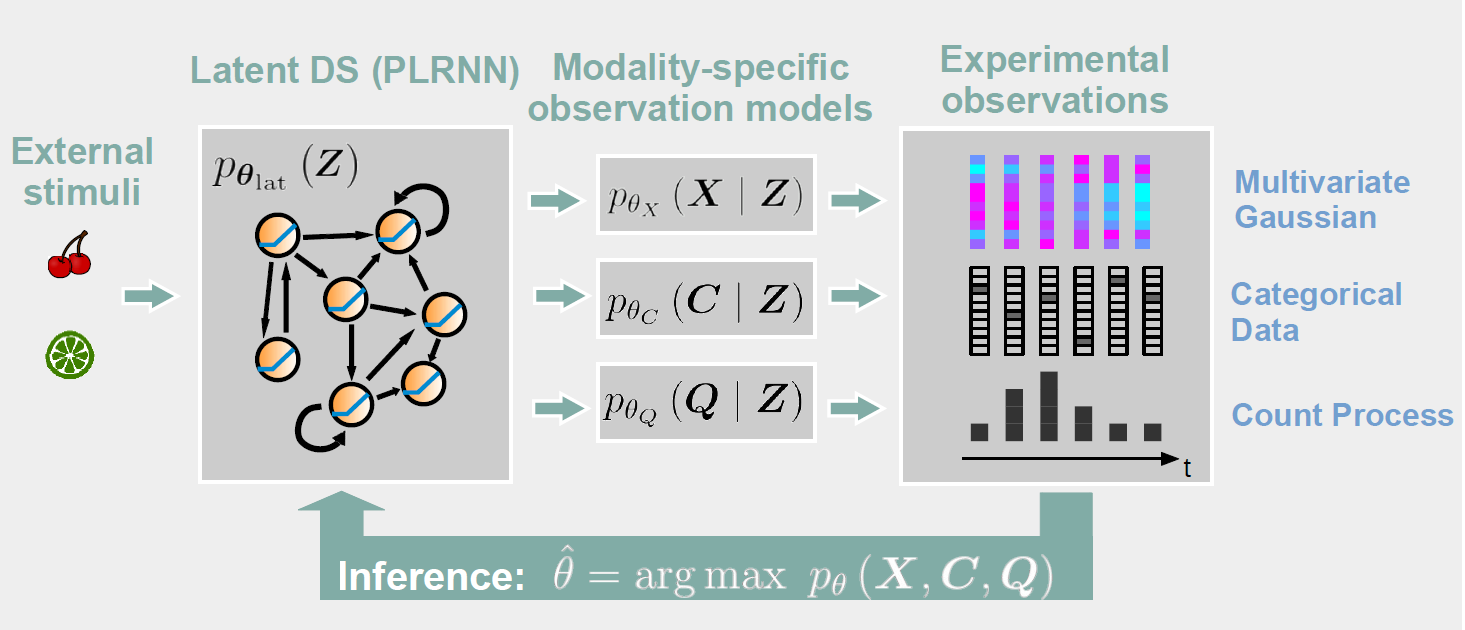}
    \end{center}
    \caption{Illustration of the multi-modal PLRNN setup: A latent DS, modelled by a PLRNN that may potentially receive external inputs, is coupled to different data modalities via modality-specific observation models.}
    \label{fig:model}
\end{figure*}

While in general the integration of multi-modal data into common predictive models has been intensely researched in recent years \cite{Sui12, Lahat15, Purdon10, Ngiam11, Srivastsava12, Turner13, Liang15,Dezfouli18,Halpern18,Bhagwat18,Antelmi18,multimodalElbo, multimodalGenModels}, so far this has hardly been a topic in the field of DS reconstruction. 
The major aim of the present work is to contribute to filling this gap. We consider the reconstruction of latent nonlinear DSs from observed time series that come from qualitatively different data domains best described by different distributional models (Fig. \ref{fig:model}). We discuss several such observation (or decoder) models from the class of generalized linear models (GLMs), but focus for most of our presentation on the case where we have both continuous Gaussian (like neural measurements) as well as distinct categorical (like behavioral information) time series, linked to the same latent RNN for approximating the DS. For training the complete system, both a novel expectation-maximization (EM) as well as a variational inference (VI) approach are developed.

\section{Related Work}
A larger body of work deals with the identification of DSs from time series data. Some of these build on physical or biological domain knowledge to set up a system of ordinary (ODE) or partial (PDE) differential equations whose parameters are to be inferred from data \cite{Gorbach17,Raissi18,Roeder19}, i.e., with the basic form of the ODE/PDE equations assumed to be known. Here, in contrast, we are interested in the case where the (exact) form of the equations is not known in advance, or where the data-generating DS is so complex (like the brain) that not all its details can be modeled, and hence we must rely on general purpose equations to approximate the underlying DS. Techniques toward this goal have been formulated both in continuous \cite{Chen2020_ODE,iakovlev2021learning} and discrete time \cite{SchmidtICLR, Zhao2020}. When using continuous-time ODE or PDE formulations \cite{chen2018,Chen2020_ODE,iakovlev2021learning}, numerical integration techniques must be used \cite{Press07}, which can be computationally quite expensive if the ODE/PDE system is stiff (like in spiking neuron models) and simple Euler or Runge-Kutta integration schemes quickly run into numerical issues \cite{Press07,Koch03}. Existing methods either aim to approximate the estimated vector field (usually obtained by differencing the time series) through (deep) neural networks \cite{Trischler16,chen2018}, RNNs \cite{Vlachas2018}, or other types of universal approximators like polynomial basis expansions \cite{Brunton16,Champion19}. Or they are directly trained on the observed time series \cite{Koppe19,SchmidtICLR,Lu17,Razaghi19,hernandez2018}, thus avoiding computation of numerical derivatives which are often unstable with large variance \cite{Chen17,Raissi18, Baydin18}.
Most of the existing techniques assume (implicitly or explicitly) the underlying set of equations to be deterministic, i.e. do not consider dynamical process noise. This is especially true for continuous-time approaches \cite{Ayed19,Champion19,Rudy19}, since stochastic DEs are even harder to deal with \cite{Risken84,Hertaeg14}. Here instead, following \cite{Durstewitz17,Koppe19,SchmidtICLR}, we assume that the generating equations are stochastic, which also helps in compensating for model misspecification \cite{Abarbanel13}. 
Fully probabilistic, generative models for DS inference have been proposed in the context of state space models and the EM algorithm \cite{Roweis02,Yu06, Durstewitz17,Koppe19,SchmidtICLR} or, more recently, based on sequential variational auto-encoders (SVAE) \cite{Krishnan15,Archer15, Zhao2020, hernandez2017,vaeComplexDynamics, Pandarinath2018}. Many of these were primarily aimed, however, at obtaining a smoothed posterior estimate $\mathrm{p}(\mathbf{Z}\mid\mathbf{X})$ of latent state trajectories $\mathbf{Z}=\{\mathbf{z}_t\}$ given observed time series $\mathbf{X}=\{\mathbf{x}_t\}$. In DS reconstruction, in contrast, we demand that the RNN, once trained, will produce \textit{simulated} data (i.e., generated from scratch from the prior $\mathrm{p}(\mathbf{Z})$) with the same temporal and geometrical structure as those from the original DS \cite{panzeri2018,SchmidtICLR}. Once successful reconstruction of the underlying DS has been achieved, it can be analyzed further with regards to its dynamical and mechanistic properties (e.g., \citealt{Brunton2019, Durstewitz, Strogatz}). For instance, computational theories in neuroscience are often cast in terms of DSs \cite{Wilson1999}, and their properties like multi-stability \cite{Durstewitz00b}, cycles \cite{Pandarinath2018} or line attractors \cite{Machens2005, Durstewitz2003} are supposed to implement functions like working memory, motor activity, decision making, or timing. Such properties and the underlying theories can be probed on DSs reconstructed from empirical data.

Regarding integration of multi-modal data, especially in the fields of computer vision \cite{Srivastsava12, multimodalElbo, multimodalGenModels} and in medical AI applications \cite{Purdon10, Liang15,Miotto16,Rajkomar18,Antelmi18} it has been demonstrated that the fusion of different data domains into a common latent representation could substantially improve predictions or reveal interesting cross-domain links \cite{Liang15}. For instance, integration of auditory and visual information into a common latent code improves speech recognition when the auditory signal is distorted, and enables cross-domain prediction \cite{Ngiam11, lee2021crossattentional}. Likewise, integration of physiological measurements like electrocardiograms or blood pressure with (categorical) entries from electronic health records (EHR) does not only enable to find important links between different data types, but also leads to better prediction of clinical outcomes \cite{Purdon10,Liang15,Rajkomar18,Antelmi18}. Most recent work focused on variational auto-encoders (VAE) and inference for multi-modal integration, assuming fully joint \cite{vedantam2018generative}, factorized \cite{Kurle2019}, mixture forms \cite{Shi2019}, or a combination of these \cite{multimodalElbo}, for the approximate posterior. Comparatively less work on multi-modal VAEs has been done, however, in the time series domain, with some exceptions especially in the area of language processing \cite{YaoHung2019,Wu2018MultimodalGM}. 

Most importantly, none of the multi-modal approaches so far was aimed at \textit{DS reconstruction} in the sense defined further above. Thus, to our best knowledge, algorithms for identifying nonlinear DSs from multiple diverse (esp. non-Gaussian) data modalities do not exist currently, although such scenarios frequently occur in the natural sciences. Here we aim to fill this gap. We will also illustrate how this allows for new types of analysis by linking different data streams to the same latent DS.

\section{Model Framework for DS Reconstruction from Multi-Modal Data}

Our complete model setup is illustrated in Fig. \ref{fig:model}. We assume that we have observed time series $\mathbf X =\{\mathbf x_t\}$ generated by some unknown DS $\text{d}\textbf{y}/\text{d}t=f(\textbf{y},t)$ sampled at discrete time points $t$ according to some output distribution $\mathrm{p}(\mathbf X\mid \mathbf Y)$, e.g. Gaussian, Poisson, or categorical. In fact, as illustrated in Fig. \ref{fig:model}, here we assume that the unknown DS is observed through several such output channels (data modalities) simultaneously, from which we would like to infer the underlying DS which is approximated by a sufficiently expressive RNN. Here we assume that all observed time series were generated by the same underlying DS and hence are naturally aligned via their common time labels (reflecting the most common scenario in the natural sciences where measurements across different modalities are taken simultaneously). 

\subsection{Generative Multi-Modal RNN Model}\label{subsec:generative model}

For our approach we build on a nonlinear state space model framework introduced previously in \cite{Durstewitz17,Koppe19}. The latent process used for approximating the unknown DS $f(\textbf{y},t)$ is modeled by a Gaussian piecewise-linear (PL) RNN of the form
\begin{align}
    \begin{split}
    \mathbf{z}_{t}\mid\mathbf{z}_{t-1} &\sim \mathcal{N}(\mathbf{A}\mathbf{z}_{t-1} + \mathbf{W}\phi(\mathbf{z}_{t-1}) + \mathbf{h} + \mathbf{F}\mathbf{s}_t, \bm{\Sigma}),\\
    \mathbf{z}_1 &\sim \mathcal{N}(\bm{\mu}_0+\mathbf{F}\mathbf{s}_1,\bm{\Sigma}),
    \end{split}
\label{encoder}
\end{align}
where $\mathbf{z}_t\in \mathbb{R}^{M\times 1}$ is the latent state vector, $\mathbf{A}\in \mathbb{R}^{M\times M}$ is diagonal with auto-regression weights $a_{mm}$, $\mathbf{W}\in \mathbb{R}^{M\times M}$ is off-diagonal (to minimize redundancy with terms in $\mathbf{A}$) with coupling weights $w_{ml}$, $m \neq l$, $\mathbf{h}\in\mathbb{R}^{M\times1}$, and $\phi(\mathbf{z}_t)=\text{max}(\mathbf{z}_t,0)$ is an element-wise ReLU transform. We also account for a time-dependent external input $\mathbf{s_t} \in \mathbb{R}^{Q\times1}$, weighted by coefficient matrix $\mathbf{F}\in \mathbb{R}^{M\times Q}$, as well as Gaussian process noise with diagonal covariance matrix $\bm{\Sigma}\in\mathbb{R}^{M\times M}$. As argued in \citet{Durstewitz17}, this model may be seen as a discretization of a neural population model (although not important for the present purpose of DS reconstruction).

One major advantage of the specific PLRNN structure in the context of DS reconstruction is that many of its dynamical properties are (analytically) tractable: Fixed points and cycles of the system can be explicitly computed \cite{SchmidtICLR}, many important types of bifurcations are comparatively well described for this class of piecewise linear maps \cite{Monfared2020-nody, SUSHKO2010}, and it can be directly translated into dynamically equivalent systems of ODEs which brings further advantages for visualization and analysis \cite{monfared2020}. This enables a detailed analysis of the learned model's behavior from a DS perspective, which is particularly important in scientific contexts where we seek to understand dynamical mechanisms beyond mere prediction of future states.
\footnote{\citet{SchmidtICLR} also discuss how to capture long-term dependencies in PLRNNs through regularization (confining the model's eigenspectrum), and how this leads different latent states to assume different functional roles in the dynamics (although this option was not used in the present work as the time series data considered here did not have very large differences in time scales).}

For inferring the latent process equations simultaneously from different data sources, the PLRNN is then connected to different observation (decoder) models that embody the specific distributional properties of the respective data domains.\footnote{On the side we note that this framework, in principle, also allows for employing more robust (non-Gaussian) distributional models for model training (e.g. \citealt{Maier2017}).} While this approach can be easily developed for almost any type of data model, especially in the flexible VI framework, in our examples we will focus on one of the most common multi-modal settings encountered in practice, namely when we have observed real-valued Gaussian time series $\mathbf{X}=\{\mathbf{x}_t\}$, $\mathbf{x}_t\in \mathbb{R}^{N\times1}$, $t=1...T$, along with multi-categorical data $\mathbf{C}=\{\mathbf{c}_t\}$, where $\mathbf{c}_t \in \{0,1\}^{K \times 1}$, $\sum_{i=1}^{K} c_{it}=1$, are binary indicator vectors. 

In this case, the latent model is connected to these two types of data domains through a Gaussian and a multi-categorical observation model, respectively, assuming that Gaussian ($\mathbf{x}_t$) and categorical ($\mathbf{c}_t$) observations are conditionally independent given latent state $\mathbf{z}_t$:
\begin{align}
    \label{decoderA}
    &\mathbf{x}_{t}\mid\mathbf{z}_{t}\sim \mathcal{N}(\mathbf{B}\phi(\mathbf{z}_t),\bm{\Gamma})~,\\
    &\mathbf{c}_t\mid\mathbf{z}_t\sim Cat(K,\bm{\pi}) ~.
    \label{decoderB}
\end{align}
The elements of the probability vector $\bm{\pi}=(\pi_1,\dots,\pi_K)^T$ are generated from the latent states $\mathbf{z}_t$ via the GLM's natural link function with regression weights $\{\bm{\beta}_i\}, i=1 \dots K-1$ (see Eq. \eqref{catprob}, Appx. \ref{suppl:carlosModels}).
In Appx. \ref{suppl:carlosModels} we also illustrate how to incorporate other exponential family models or mixtures thereof into our framework, but this Gaussian $+$ categorical setting will suffice to convey the basic principles and features of our algorithm. 
We call the resulting model, Eq.~(\mbox{\ref{encoder}-\ref{decoderB}}), the \textit{multi-modal} PLRNN (mmPLRNN) and the model defined by Eq.~(\mbox{\ref{encoder}-\ref{decoderA}}) the \textit{uni-modal} PLRNN (uniPLRNN). 

We would like to infer the mmPLRNN parameters ${\bm{\theta}=\{\bm{\mu}_0,\mathbf{A},\mathbf{W},\mathbf{F},\mathbf{h},\mathbf{B},\{\bm{\beta}_i\},\bm{\Gamma},\bm{\Sigma}\}}$ and latent states ${\mathbf{Z}=\{\mathbf{z}_t\}}$ from the set of observations $\{\mathbf{X},\mathbf{C}\}$ by maximizing the likelihood
\begin{align}
\begin{split}
\mathrm{p}_{\bm{\theta}}(\mathbf{X},\mathbf{C})=&\int_{\mathbf{Z}}\mathrm{p}_{\bm{\theta}}(\mathbf{z}_1) \prod_{t=2}^{T}\mathrm{p}_{\bm{\theta}}(\mathbf{z}_t\mid\mathbf{z}_{t-1})\\ &\prod_{t=1}^{T}\mathrm{p}_{\bm{\theta}}(\mathbf{x}_t\mid\mathbf{z}_t)
\prod_{t=1}^{T}\mathrm{p}_{\bm{\theta}}(\mathbf{c}_t\mid\mathbf{z}_t) \mathrm{d}\mathbf{Z},
\end{split}
\label{JP}
\end{align}
where we have used the conditional independence of Gaussian and categorical observations. Observations missing in one or both channels at any time point $t$ may simply be dropped from the likelihood Eq. \eqref{JP} while training. 
Since the log-likelihood is intractable for our model, we replace it by the evidence lower bound (ELBO, \citealt{Kingma13, Blei2017}), which in our case is
\begin{align}
\small
    \begin{split}
    \mathcal{L}(\bm{\theta},q):&= E_q\left[\log \mathrm{p}_{\bm{\theta}} (\mathbf{X},\mathbf{C},\mathbf{Z}) \right] + H\left[ \mathrm{q}(\mathbf{Z}\mid\mathbf{X},\mathbf{C}) \right] \\
    &= \log \mathrm{p}_{\bm{\theta}} (\mathbf{X},\mathbf{C}) - \mathrm{KL}\left[\mathrm{q}(\mathbf{Z}\mid\mathbf{X},\mathbf{C})\parallel \mathrm{p}_{\bm{\theta}}(\mathbf{Z}\mid\mathbf{X},\mathbf{C})\right] \\
    &\leq \log \mathrm{p}_{\bm{\theta}}(\mathbf{X},\mathbf{C}),
    \end{split}
    \label{ELBO}
\end{align}
where $\mathrm{q}(\mathbf{Z}\mid\mathbf{X},\mathbf{C})$ is a proposal or variational density.

In the next two sections we will introduce both an efficient EM as well as a VI algorithm for maximizing the ELBO.\footnote{Note that this is a very different approach to dynamics than reservoir computing (RC; e.g. \citealt{Pathak2018}): While RC essentially learns to map a rich but rather fixed dynamical repertoire to the observations, here we aim to learn the underlying DS itself within the (tractable) PLRNN equations. Hence, fixing the latent and just training the observation model cannot in general work for the PLRNN (as we also explicitly checked just as a control).}

\subsection{Expectation Maximization (EM) for Model Training}
\label{sec:EM}
It has been shown previously \cite{Durstewitz17,Koppe19} that the piecewise-linear structure of model Eq.~(\ref{encoder}) can be efficiently exploited in EM by a fixed-point iteration algorithm and partly analytical derivation of expectations, on which we will build here.
\paragraph{State Estimation}
In the E-step we assume, similar to a Laplace-Gaussian approximation, that the expectation value $\mathrm{E}[\mathbf{Z}\mid\mathbf{X},\mathbf{C}]$ is reasonably well approximated by the mode, and solve the following maximization problem:
\begin{align}
\small
\begin{split}
    \text{E}&[\mathbf{Z}\mid\mathbf{X},\mathbf{C}]\approx \mathbf{Z}^{\text{max}} 
    :=\argmax_\mathbf{Z}\left[ \log \mathrm{p}_{\bm{\theta}}(\mathbf{Z}\mid\mathbf{X},\mathbf{C})\right]\\
    &=\argmax_\mathbf{Z} \left[\log \mathrm{p}_{\bm{\theta}}(\mathbf{C}\mid\mathbf{Z})+ \log \mathrm{p}_{\bm{\theta}}(\mathbf{X}\mid\mathbf{Z})
     + \log \mathrm{p}_{\bm{\theta}}(\mathbf{Z})\right. \\ 
     & ~~~~~~~~~~~~~~~~~~~~~~  - \left.\log \mathrm{p}_{\bm{\theta}}(\mathbf{X},\mathbf{C})\right]~\\
    & = \argmax_\mathbf{Z}\left[\log \mathrm{p}_{\bm{\theta}}(\mathbf{C}\mid\mathbf{Z}) +  \log \mathrm{p}_{\bm{\theta}}(\mathbf{X}\mid\mathbf{Z}) +  \log \mathrm{p}_{\bm{\theta}}(\mathbf{Z})\right]~.
    \end{split}
\label{GaussApp}
\end{align}
The covariance matrix of $\mathrm{p}_{\bm{\theta}}(\mathbf{Z}\mid\mathbf{X},\mathbf{C})$ is then approximated by the negative inverse Hessian $(-\mathbf{H}^{\text{max}})^{-1}$ around this maximizer, based on which all state expectations $\mathrm{E}[\mathbf{z}]$, $\mathrm{E}[\mathbf{z}\mathbf{z}^T]$, $\mathrm{E}[\phi(\mathbf{z})]$, $\mathrm{E}[\mathbf{z}\phi(\mathbf{z})^T]$ and $\mathrm{E}[\phi(\mathbf{z})\phi(\mathbf{z})^T]$ needed for the M-step can be computed (semi-)analytically for the PLRNN \cite{Durstewitz17,Koppe19}. 

In the original formulation of the PLRNN algorithm \cite{Durstewitz17,Koppe19}, criterion Eq.~(\ref{GaussApp}) was piecewise quadratic (owing to the piecewise linear ReLU activation) and could be addressed by an efficient fixed-point-iteration algorithm. Due to the non-Gaussian terms in $\mathrm{p}(\mathbf{C}\mid\mathbf{Z})$, this is no longer the case. But for any exponential family function in the decoder, Eq.~(\ref{GaussApp}) will remain piecewise concave (within each orthant) and can be addressed by an efficient Newton-Raphson (NR) scheme (see Appx. \ref{suppl:Newton-Raphson} for details). 
\paragraph{Parameter Estimation}
For parameter estimation (M-step) we assume we have all relevant moments of  
$\mathrm{q}(\mathbf{Z}\mid\mathbf{X,C})$ from the E-step and, based on this, solve the maximization problem $\bm{\theta^{\ast}}:=\text{arg max}_{\bm{\theta}}\mathcal{L}(\bm{\theta},\mathrm{q}^{\ast})$. In the original PLRNN state space model defined by Eq.~(\ref{encoder} - \ref{decoderA}) one can solve this analytically and quickly in one step as a linear regression problem given all expectations in $\mathbf{Z}$. 
This is still true here for all parameters that define the latent state prior model $\mathrm{p}_{\bm{\theta}_{\text{lat}}}(\mathbf{Z})$ (Eq.~(\ref{encoder})), $\bm{\theta}_{\text{lat}}=\{\bm{\mu}_0,\mathbf{A},\mathbf{W},\mathbf{F},\mathbf{h},\bm{\Sigma}\}$, and those occurring within the Gaussian observation model ${\mathrm{p}_{\bm{\theta}_X}(\mathbf{X}\mid\mathbf{Z})}$, $\bm{\theta}_X=\{\mathbf{B},\bm{\Sigma}\}$. 
However, the terms in $\mathrm{E}[\log \mathrm{p}(\mathbf{C}\mid\mathbf{Z})]$ are a bit more tricky. 
To separate model parameters $\boldsymbol{\theta}$ from expectations in states $\mathbf{z}_t$, we therefore introduce another lower bound into the log-likelihood using Jensen’s inequality ($*$) that makes the problem tractable: 
\begin{align}
\small
\begin{split}
\text{E}&[\log \mathrm{p}_{\bm{\theta}}(\mathbf{C}\mid\mathbf{Z})] = \\
&\sum_{t=1}^{T}\text{E}\left[\sum_{i=1}^{K-1} c_{it}\bm{\beta}_{i}^T\mathbf{z}_t\right]  - \sum_{t=1}^{T}\text{E}\left[\log \left(1+\sum_{j=1}^{K-1}  \exp(\bm{\beta}_j^T\mathbf{z}_t) \right)\right]\\
&\overset{(*)}{\geq}\sum_{t=1}^{T}\sum_{i=1}^{K-1} c_{it}\bm{\beta}_{i}^T\text{E}[\mathbf{z}_t] - \sum_{t=1}^{T}\log \left(1+\sum_{j=1}^{K-1} \text{E}[ \exp(\bm{\beta}_j^T\mathbf{z}_t)]\right)~.
\end{split}
\label{firstsimp}
\end{align}

Further noting that states $\mathbf{z}_t$ are conditionally Gaussian, we can use the moment-generating function of the Gaussian (see also \citealt{Brown03}) to reshape Eq.~(\ref{firstsimp}) as
\begin{align}
\small
    \begin{split}
        f(\bm{\beta})&:=\sum_{t=1}^{T} \sum_{i=1}^{K-1} c_{it}\bm{\beta}_{i}^T\text{E}[\mathbf{z}_t] \\
        &-\sum_{t=1}^{T}\log \left[1+\sum_{j=1}^{K-1} \exp\left(\bm{\beta}_j^T\text{E}[\mathbf{z}_t]+\frac{\bm{\beta}_j^T\text{Cov}(\mathbf{z}_t)\bm{\beta}_j}{2}\right)\right]~.
        \label{gtheta}
    \end{split}
\end{align}
This is a concave function again in parameters $\bm{\beta}$ that only requires expectations $\text{E}[\mathbf{z}_t]$ and $\text{E}[\mathbf{z}_t\mathbf{z}_t^T]$ 
from the E-step, which can hence be solved quickly and efficiently by NR iterations (see Appx. \ref{suppl:Newton-Raphson}).

Since all exponential family distributions, as well as sums of exponential family models, have concave log-likelihoods \cite{Fahrmeier2001}, one can always employ the NR scheme for the E- and M-steps as in Eq. \eqref{update} and \eqref{fullNR}, as long as the distributional parameters are connected to the latent states through the natural link function. This makes the above algorithm more generally applicable (beyond just Gaussian and categorical observations). For more details on training see Appx. ~\ref{suppl:Newton-Raphson} \& Appx. ~\ref{suppl:stepwiseTraining}, \textit{Stepwise training protocol}.

\begin{figure*}[!th]
    \centering
    \includegraphics[width=1.0\textwidth]{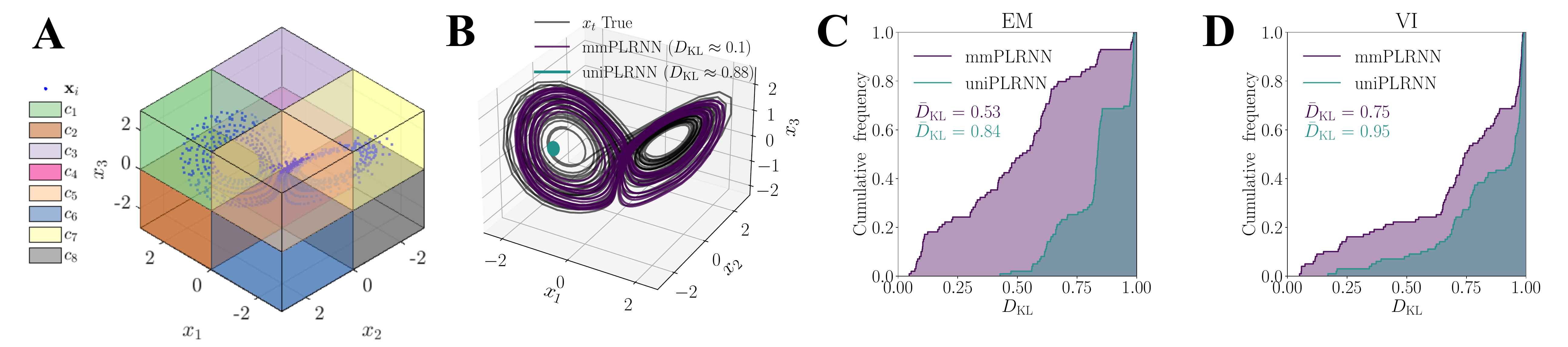}
    \caption{Improving DS reconstruction with multi-modal data when continuous observations are too noisy. A) Experimental setup with Gaussian and categorical information.
    B) Example of successful DS reconstruction with multi-modal (purple) but not uni-modal (cyan) PLRNN. Black trajectory = ground truth.
    C) Cumulative performance histograms ($n=100$ runs) in terms of normalized Kullback-Leibler divergence $D_\text{KL}/D_\text{KL}^\text{max}$  between true and generated attractor geometries for uni- vs. multi-modal PLRNN produced by the EM algorithm. $\Bar{D}_\text{KL}$ indicates the median. 
    D) Same for models trained through VI.}
    \label{NoisyLorenz}
\end{figure*}

\subsection{Variational Inference for Model Training}
\label{sec:VI}
The EM algorithm for PLRNN inference has been shown to efficiently work with small data sets \cite{Koppe19}, but it lacks flexibility (other than exponential family distributions may be harder to accommodate). 
An alternative way to optimize expression (\ref{ELBO}) is VI, whereby $\mathrm{q}_{\bm{\zeta}} \left(\mathbf{Z}\mid\mathbf{X},\mathbf{C}\right)$ becomes a parameterized family of variational densities for approximating the true posterior. 
We model the approximate posterior by a multivariate Gaussian, 
\begin{equation}
\mathrm{q}_{\boldsymbol\zeta}\left( \mathbf{Z} \mid \mathbf{X}, \mathbf{C} \right) = 
\mathcal{N}\left(\boldsymbol{\mu}_Z(\mathbf{X}, \mathbf{C}), \mathbf{\Lambda}_Z(\mathbf{X},
\mathbf{C})\right)~,
\label{eq:approximatePosterior}
\end{equation}
where the mean $\boldsymbol{\mu}_Z(\mathbf{X}, \mathbf{C}) \in \mathbb{R}^{M T \times 1}$ and covariance matrix $\mathbf{\Lambda}_Z(\mathbf{X}, \mathbf{C}) \in \mathbb{R}^{M T \times M T}$ are parameterized by neural networks with parameters $\boldsymbol\zeta=\{\boldsymbol\zeta_\mu,\boldsymbol\zeta_\Lambda\}$. Specifically, instead of parameterizing a full covariance $\mathbf{\Lambda}$ directly, we follow \cite{Archer15} and parameterize the $M \times M$ on- and off-diagonal blocks of the Hessian $\mathbf{H} = -\mathbf{\Lambda}_Z^{-1}$ by neural networks, exploiting its block tri-diagonal structure owing to the Markovian latent model assumptions (see Appx. \ref{suppl:paramterization} for more details).

Joint optimization of variational ($\boldsymbol\zeta$) and generative ($\boldsymbol\theta$) model parameters is performed via Stochastic Gradient Variational Bayes (SGVB) using the re-parameterization trick for the model's random variables \cite{Kingma13,Rezende14}. We chose 
Adam \cite{Adam} with learning rate $\alpha=10^{-3}$.

All code produced here is available at \url{github.com/DurstewitzLab/mmPLRNN}.

\section{Results}\label{results}
We first evaluate the algorithm’s ability to combine information from different, distinct data streams into a common latent nonlinear DS model on two purpose-designed ground truth systems. For these we produce both continuous Gaussian and categorical information from the Lorenz ODE system within its chaotic regime \cite{Lorenz1963}, a popular benchmark system for testing DS reconstruction. We then probe our algorithm on experimental data consisting of simultaneous functional Magnetic Resonance Imaging (fMRI) recordings of different brain regions and (categorical) behavioral data from healthy subjects performing a working memory task \cite{Koppe2014}. Detailed information on hyper-parameter settings for all methods and experiments is collected in Appx. \ref{suppl:Hypers}.

\begin{figure*}[t!]
    \centering
    \includegraphics[width=0.93\textwidth]{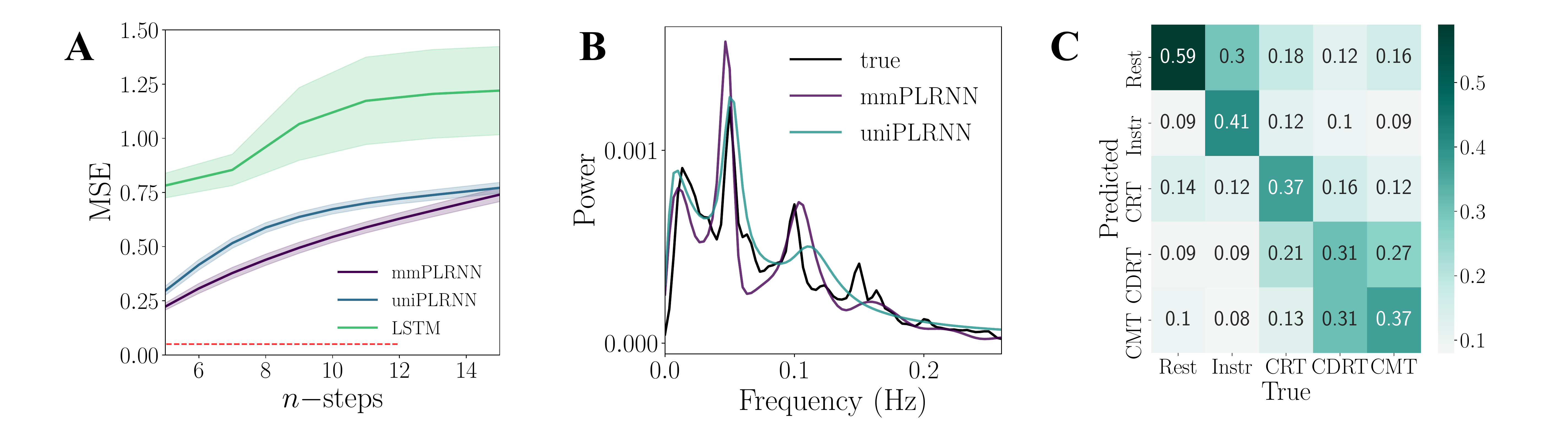}
    \caption{mmPLRNN trained on simultaneous BOLD recordings and categorical data from fMRI experiments. A) MSE for $n$-step ahead predictions for uni- vs. mmPLRNN, and for LSTMs trained with Adam and learning rate $\alpha=0.005$. fMRI sampling rate was $1/3 $ Hz, so one step corresponds to $3$ s. Error shadings indicate SEM. For the indicated time steps (red dots), the difference between uni- and mmPLRNN is significant ($p<.05$) in Tukey post-hoc tests. Note that for a chaotic system, multimodal gains in predictability are expected to vanish for larger time steps due to the exponential divergence of trajectories. B) Example reproduction of power spectra.  C) Confusion matrix of predicted vs. true class labels on test sets. Base rates of classes were $0.32$ (Rest), $0.125$ (Instr), $0.185$ (CRT), $0.185$ (CDRT), $0.185$ (CMT). 
    } 
    \label{ExpEval}
\end{figure*}

\subsection{Benchmarks: Noisy or Incomplete Lorenz System with Gaussian and Categorical Observations}\label{subsec:noisyLorenz}
The 3D-Lorenz system is defined by the set of Eqs. (\ref{Lorenz}) (see Appx. \ref{suppl:Lorenz}) where we have added a Gaussian dynamical (process) noise term $\mathrm{d}\bm{\epsilon}(t)\sim\mathcal{N}(0,0.0025\times \mathrm{d}t \mathbf{I})$ when integrating the equations, making this a system of stochastic differential equations. State trajectories $\mathbf{x}_t=(\mathrm{x}_1,\mathrm{x}_2,\mathrm{x}_3)^{T}$ were generated from this system (Fig.~\ref{NoisyLorenz}A) using 
fourth-order Runge-Kutta numerical integration \cite{Press07}. Generated trajectories were further standardized (centered and scaled to unit variance on each dimension) and blurred by adding a relatively large amount of Gaussian observation noise $\bm{\eta}_t\sim\mathcal{N}(0,0.1 \times \mathbf{I})$, strongly degrading the information about the underlying DS that could be obtained from the continuous Gaussian observations alone. This emulates a natural scenario where one information channel on its own may be too noisy to enable identification of the underlying system. 
In addition to these Gaussian observations, we provide categorical information about the system’s dynamics by indicating in which of the eight orthants the system’s current state is in (Fig.~\ref{NoisyLorenz}A), i.e., in the form of an 8-dimensional indicator vector $\mathbf{c}_t=(c_{1t},\dots,c_{8t})^T$, where $c_{it}=1$ for the $(I[x_{1t}>0]2^0+I[x_{2t}>0]2^1+I[x_{3t}>0]2^2+1)^{th}$ bit and $0$ otherwise. The mmPLRNN algorithm had access to only one relatively short time series $\{\mathbf{x}_t,\mathbf{c}_t\}$, $t=1...T$, of length $T=1000$ to infer the underlying DS, using $M=15$ latent states (see Appx. \ref{suppl:Hypers}). 

To evaluate the quality of DS reconstruction, new trajectories were sampled from the once trained generative model $\mathrm{p}_{\bm{\theta}}(\mathbf{X},\mathbf{C},\mathbf{Z})$, i.e. new latent state trajectories were first drawn from the model prior $\mathbf{Z}\sim\mathrm{p}_{\bm{\theta}_\mathrm{lat}}(\mathbf{Z})$ defined by the latent process Eq.~(\ref{encoder}), from which time series observations $\mathbf{X}\sim\mathrm{p}_{\bm{\theta}_{X}}(\mathbf{X}\mid\mathbf{Z})$ and $\mathbf{C}\sim\mathrm{p}_{\bm{\theta}_{C}}(\mathbf{C}\mid\mathbf{Z})$ were produced according to the learned observation models. Fig.~\ref{NoisyLorenz}B provides an example of successful reconstruction of the Lorenz system, i.e. where the mmPLRNN’s intrinsic dynamics captures well the butterfly-wing structure of the chaotic Lorenz attractor.

To quantify reconstruction quality, we used a previously introduced Kullback-Leibler measure for the agreement between true, $\hat{\mathrm{p}}_{\text{true}}(\mathbf{x})$, and model-generated, $\hat{\mathrm{p}}_{\text{gen}}(\mathbf{x}\mid\mathbf{z})$, \textit{attractor geometries} \cite{Koppe19} (see Eq. (\ref{KLdis}) in Appx. \ref{suppl:kullback}). Importantly, this measure assesses the agreement across space, not time: 
As pointed out in \cite{Woods10,Koppe19}, trajectories from chaotic systems not started from exactly the same initial condition exponentially diverge, potentially leading to ultimately highly dissimilar time series with large MSE deviation, even though they may have been generated by the very same DS (see Fig. 2 in \citealt{Koppe19}). In contrast, (time-) invariant properties like 
attractor geometries should be preserved. As shown in Fig. \ref{NoisyLorenz}C-D, attractor reconstructions as assessed by this measure strongly improve when the algorithm has access to the categorial information on top of the Gaussian time series, in contrast to when only the latter were available. This was true regardless of whether the mmPLRNN was trained by EM or VI, although for EM the improvement was 
more pronounced and reconstructions were better on average (this advantage of EM disappeared, however, if VI was allowed access to longer training sequences, $T=2500$: $\Bar{D}_\text{KL} \approx 0.49 \ (n=40)$, see also Fig. \ref{fig:noiseLevels}).  
This demonstrates that the mmPLRNN can strongly enhance DS identification by supplementing the highly noisy trajectory information by categorical data, even though, and importantly, these do not provide \textit{quantitative} information about the dynamics (in particular, no information about attractor geometry or topology). Of course, as noise levels become very low, information from a continuous-Gaussian channel alone may enable sufficiently good reconstructions (as further explored in Fig. \ref{fig:noiseLevels}).

As a second test case, we studied whether additional categorical information could also help to identify the chaotic Lorenz system when one of its dynamical variables ($x_2$) was missing from the observations, i.e., only $\mathbf{x}^{\mathrm{red}}_t=(x_{1t},x_{3t})^T$ was provided for training. This is indeed the case, as reported in Appx. \ref{sec:missDim}, Fig. \ref{fig:ReducedLorenz}.

\subsection{Empirical example: DS inference from Neurophysiological and Task Label Data}\label{ExpData}
For probing the mmPLRNN on real data, we chose a data set consisting of fMRI recordings (which assess the so-called Blood-Oxygenation-Level-Dependent, BOLD, signal) taken from human subjects while they performed simple working memory and control tasks. The details of the experimental setup are not overly important here and are given in \citet{Koppe2014} and briefly summarized in Appx. \ref{suppl:fmriDetails}. $N=20$ brain regions (from $l = 21$ subjects, see Appx. \ref{suppl:fmriDetails}) were selected for analysis, yielding continuous time series data $\mathbf{X}=\{\mathbf{x}_t\}$, $\mathbf{x}_t=(x_{1,t},\dots,x_{20,t})^T$ for each subject. Of note, BOLD time series were relatively short ($T=360$) and hence extracting reasonable dynamics from \textit{single subjects} is quite challenging. In fact, for this type of very sparse data only the more efficient EM algorithm tended to produce 
successful reconstructions, and hence only these are reported here. 
As categorical data we defined the five major task stages subjects underwent during the experiment ('Rest', 'Instruction', 'Choice Reaction Task', 'Continuous Delayed Response Task', 'Continuous Matching Task'), and endowed each time step with a categorical label $\mathbf{c}_t \in \{0,1\}^5$ accordingly.\footnote{
We stress that these different task stages did \textit{not} differ in the type of presented stimuli or required responses, i.e. in their 'external inputs', but rather tapped into different cognitive processes.
} 

\begin{figure*}[!ht]
    \centering
    \includegraphics[width=0.9\textwidth]{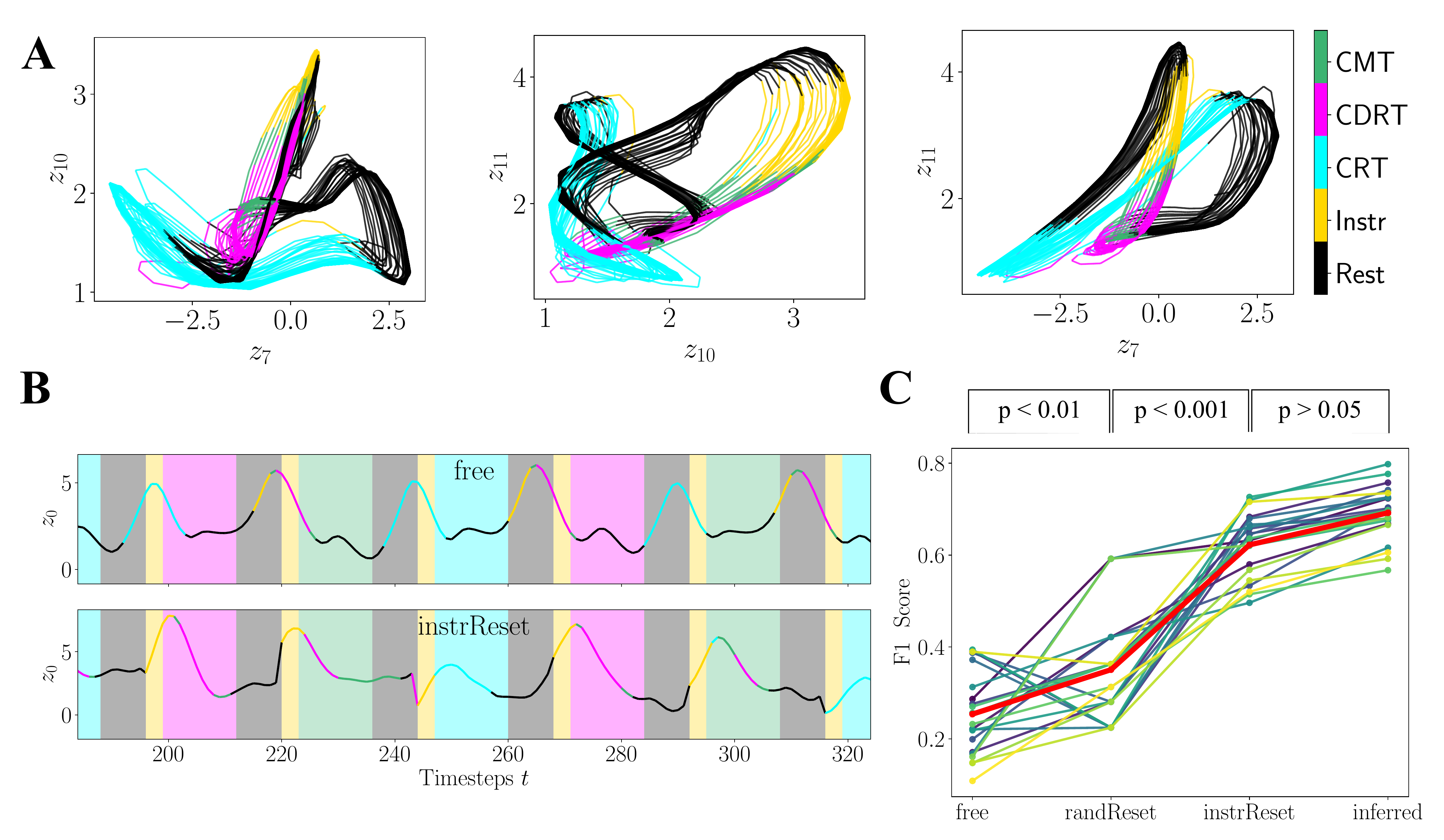}
    \caption{Neural dynamics underlying task stages. A) Association between predicted class labels (color coding) and learned BOLD dynamics. Shown are 2d subspaces of an mmPLRNN's generated state space. Subspaces chosen for display were selected according to Fisher's discriminant criterion. B) Top: Freely generated latent activity (initialized only once with the inferred state at the beginning of the experiment), color-coded according to task stage as in \textbf{A}. True task stages indicated by the pale background colors. Bottom: Same, but with generated latents reset to inferred values upon each new instruction phase. 
    C) Summary statistics across $l=21$ subjects, comparing task stage decoding from simulated latent activity initialized only at experiment onset (``free''), 
    simulated activity reset at $15$ random time bins (``randReset''), 
    simulated activity reset at the $15$ instruction onsets (``instrReset''), 
    and fully inferred latent states (``inferred'', i.e. the conditional means $\mathrm{E}[\mathbf{Z}\mid\mathbf{X}]$). 
    Red line = group means. Repeated measures ANOVA 
    ($F\approx138.56, p < 10^{-5}$) with Tukey posthoc-tests as indicated.} 
    \label{Fig4}
\end{figure*}

As in the Lorenz benchmark setups, we first studied whether including categorical information would help the algorithm to produce better reconstructions and predictions as compared to when only BOLD time series were provided. Fig. \ref{ExpEval}A shows the ahead-prediction error for $n \in \{5 \dots 15\}$ 
future time steps, and for both a uni-modal PLRNN, trained only on the BOLD signals, and the mmPLRNN which consumed class labels in addition (both trained with $M=20$ states; 
for comparison, also predictions produced by a LSTM with the same number of latent states are shown). There is a consistent and overall highly significant (repeated ANOVA, $p<.001$) 
increase in performance (up to $26\%$) for the mmPLRNN across all prediction time steps. The mmPLRNN also produced on average significantly better geometrical reconstructions as revealed by our Kullback-Leibler criterion (unnormalized $\langle D_{mm} \rangle \approx 1.09 \pm 0.03 < \langle D_{uni} \rangle \approx 2.56 \pm 0.03$, $p<10^{-4}$, Wilcoxon ranksum test). Thus, the additional categorical information indeed significantly helped to reconstruct the underlying system. 
The example of true and reconstructed power spectra in Fig. \ref{ExpEval}B 
furthermore confirms that the mmPLRNN has learned to generate (i.e., freely simulate) time series which exhibit the same major temporal signatures (peaks in the spectrum) as the real data. 
Hence, also for this empirical example the mmPLRNN seems to exploit both data modalities to achieve the best reconstruction.
This is an important and non-trivial result, as it confirms that even in this empirical scenario purely categorical information may help in guiding the algorithm toward better approximations of the underlying neural dynamics.

Furthermore, we tested cross-modal inference, i.e. whether the trained mmPLRNN would be able to predict class labels from BOLD signals alone on left-out test data. While here this mainly serves to examine whether cross-modal links have been built within the model's latent space, it is also a 
relevant application case. 
Specifically, we ran a cross-validation protocol where each $20\%$ segment of the time series was left out in turn for training, 
and 
unseen class labels were predicted on these left-out test sets (see Appx. \ref{suppl:fmriDetails} for details). 
Fig.~\ref{ExpEval}C summarizes the agreement between true and predicted behavioral class labels across all test sets from successful training runs (see Appx. \ref{suppl:fmriDetails}) 
in a confusion matrix. These results were 
on par with those produced by various classifiers trained \textit{directly} on the same $20$-dimensional BOLD signal vectors and associated class labels as used for the mmPLRNN 
(multi-class F1 scores for logistic regression: $\approx 0.47$, linear discriminant analysis: $\approx 0.48$, support vector machines (RBF kernel, $C=1$): $\approx 0.47$, mmPLRNN: $\approx 0.45$). 
This confirms that the mmPLRNN has learned the connections between the two data modalities within its latent space in an about optimal manner, i.e., without much loss of information as judged by this 
comparison. 

Of course, in practice we would not use the mmPLRNN merely for cross-modal prediction. Rather, the main purpose of this approach is that we can now harvest the trained model and common latent representation to investigate the \textit{dynamical mechanisms} of the observed BOLD signals and 
cross-modal links. In general, properly trained mmPLRNNs exhibited complex cycles (often chaotic oscillators, Fig. \ref{Fig4}A) as their limiting behavior (i.e., attractors) that mimic the temporal structure of the BOLD signal. For the example in Fig. \ref{Fig4}A we 
analytically computed the PLRNN's maximum Lyapunov exponent as $\lambda_\text{max} \approx 0.009$ (
attesting its chaotic nature, see Appx. \ref{sec:LyapExp}). As the example shows, different task stages seem to be associated with different subcycles or phases of the chaotic oscillator. Across all subjects, \textit{freely running} mmPLRNN oscillators\footnote{By this we mean $\mathrm{E}[\mathbf{Z}]$ computed by forward-iterating Eq. \ref{encoder} in time from $\bm{\mu}_0$ as inferred from the very first time step.} initially (at the start of an experiment) predicted task stages quite well, but then started to run out of phase with the experimental stages (significantly better agreement with true class labels in $1^{st}$ (F1$\approx0.38$) compared to $2^{nd}$ (F1$\approx0.17$) and $3^{rd}$ (F1$\approx0.16$) thirds of time series; $F\approx30.68$, $p < 10^{-5}$). 
This is expected since, unlike the real experimental subjects who were updated with each new instruction phase, the freely simulated mmPLRNN does not receive any task-related information but simply follows its internal dynamics. Indeed, resetting the intrinsic PLRNN oscillators at the beginning of each instruction phase to the inferred latent state (i.e., posterior estimate $\mathrm{E}[\mathbf{z}_t\mid\mathbf{X}]$) significantly improved the alignment with the experimental task stages (Fig. \ref{Fig4}B; Fig. \ref{Fig4}C, ``instrReset''); in particular, significantly more so than just resetting the PLRNN oscillators to inferred values at arbitrary (but similarly spaced) time points throughout the experiment (Fig. \ref{Fig4}C, ``randReset''). In contrast, replacing \textit{all} simulated latent states by inferred values did not yield a significant improvement in task alignment anymore (Fig. \ref{Fig4}C, ``inferred''). Hence, the mmPLRNN explains the links between BOLD activity and task stages through a complex oscillator that is differently initialized upon each distinct instruction phase. 

While multivariate classifiers have long been used for reading out sensory or cognitive representations from fMRI activity \cite{Haynes2006, HAYNES2015}, methods like the mmPLRNN therefore enable to reveal much more specifically, in terms of DS mechanisms, how BOLD dynamics and mental processes are linked. Neural oscillations, in particular, play a huge role in cognition and memory formation \cite{Buzski2006}. The functional significance of slower oscillations as detectable with fMRI is as yet unclear \cite{DREW2020782, Lewis2016}, however, where the present methods may help to improve our understanding. Exploring these possibilities further for other fMRI task conditions or subject groups (patients) would be an interesting future direction.

\section{Conclusions}\label{Disscussion}
In this work we introduced a new algorithm for nonlinear DS reconstruction, the mmPLRNN, that draws on several data channels described by different distributional models. By DS reconstruction here we meant that the trained system approximates the true underlying DS in a generative sense, i.e. such that after training trajectories drawn (simulated) from the latent process would produce the same state space behavior and invariant properties (like attractor geometries) as the observed system. Although various approaches toward this goal have been introduced before 
\cite{Brunton16,Champion19,Duncker19,Lu17,Durstewitz17,Koppe19,Razaghi19}, 
to our knowledge the mmPLRNN is the first such system that can integrate different types of data modalities for this purpose. We developed both an EM- and a VI-based variant of the basic algorithm, and demonstrated that the mmPLRNN will use categorical (or other, see Appx. \ref{suppl:carlosModels}) information to fill in for information too noisy or missing in the Gaussian channel, i.e. will effectively combine the different information sources to infer the underlying DS. 
We also showcased the mmPLRNN on a neuroscientific dataset consisting of simultaneous fMRI recordings and behavioral task labels, and showed how it could be used to gain insight into the neural dynamics underlying BOLD-class label associations. 

Both the EM- and VI-based algorithms have their own advantages and drawbacks: VI scales better with problem size than EM, as it can be efficiently trained through gradient descent using the reparameterization trick \cite{Kingma13,Rezende14, Krishnan15,Archer15}. 
It is also more flexible as it can more easily accommodate a larger variety of distributional models.  
For the EM-based mmPLRNN, on the other hand, although the steps outlined here for categorical data 
are fairly easy to extend to other exponential family models,
more complex distributional models would require additional thought and possibly hand-crafted solutions. On the upside, the EM algorithm can more efficiently deal with smaller scale problems as often encountered in physical or biological experiments (like the fMRI data studied here), and provides higher accuracy estimates that may be preferable in scientific or medical settings. This is because for the PLRNN our EM can -- given $\mathrm{p}(\mathbf{Z}\mid\mathbf{X,C})$ -- compute many of the parameters and expectations analytically and, in contrast to VI, is second-order even in its numerical parts (see Appx. \ref{suppl:Newton-Raphson}). An interesting question for future investigation is how much information about an underlying DS can be recovered \textit{solely} from non-continuous and non-Gaussian data, like categorical or count series.

\section{Acknowledgements}
This work was funded by grants from the German Research Foundation (DFG) within the CRC TRR-265 (A06 \& B08) to DD and GK, and through Du 354/10-1 (DFG) to DD.






\medskip

\small
\bibliography{bibliography}

\begin{thebibliography}{95}
\providecommand{\natexlab}[1]{#1}
\providecommand{\url}[1]{\texttt{#1}}
\expandafter\ifx\csname urlstyle\endcsname\relax
  \providecommand{\doi}[1]{doi: #1}\else
  \providecommand{\doi}{doi: \begingroup \urlstyle{rm}\Url}\fi

\bibitem[Abarbanel(2013)]{Abarbanel13}
Abarbanel, H.
\newblock \emph{Predicting the Future: Completing Models of Observed Complex
  Systems (Understanding Complex Systems)}.
\newblock Springer, 2013.
\newblock ISBN 978-1-4614-7218-6.

\bibitem[Antelmi et~al.(2018)Antelmi, Ayache, Robert, and Lorenzi]{Antelmi18}
Antelmi, L., Ayache, N., Robert, P., and Lorenzi, M.
\newblock Multi-channel stochastic variational inference for the joint analysis
  of heterogeneous biomedical data in alzheimer's disease.
\newblock In Stoyanov, D., Taylor, Z., Kia, S.~M., Oguz, I., Reyes, M., Martel,
  A.~L., Maier{-}Hein, L., Marquand, A.~F., Duchesnay, E., L{\"{o}}fstedt, T.,
  Landman, B.~A., Cardoso, M.~J., Silva, C.~A., Pereira, S., and Meier, R.
  (eds.), \emph{Understanding and Interpreting Machine Learning in Medical
  Image Computing Applications - First International Workshops {MLCN} 2018,
  {DLF} 2018, and iMIMIC 2018, Held in Conjunction with {MICCAI} 2018, Granada,
  Spain, September 16-20, 2018, Proceedings}, volume 11038 of \emph{Lecture
  Notes in Computer Science}, pp.\  15--23. Springer, 2018.
\newblock \doi{10.1007/978-3-030-02628-8\_2}.
\newblock URL \url{https://doi.org/10.1007/978-3-030-02628-8\_2}.

\bibitem[Archer et~al.(2015)Archer, Park, Buesing, Cunningham, and
  Paninski]{Archer15}
Archer, E., Park, I.~M., Buesing, L., Cunningham, J., and Paninski, L.
\newblock Black box variational inference for state space models.
\newblock \emph{ArXiv}, 2015.

\bibitem[Ayed et~al.(2019)Ayed, de~Bézenac, Pajot, Brajard, and
  Gallinari]{Ayed19}
Ayed, I., de~Bézenac, E., Pajot, A., Brajard, J., and Gallinari, P.
\newblock Learning dynamical systems from partial observations.
\newblock \emph{arXiv}, 2019.

\bibitem[Baydin et~al.(2018)Baydin, Pearlmutter, Radul, and Siskind]{Baydin18}
Baydin, A.~G., Pearlmutter, B.~A., Radul, A.~A., and Siskind, J.~M.
\newblock Automatic differentiation in machine learning: a survey.
\newblock \emph{Journal of machine learning research}, 2018.

\bibitem[Bhagwat et~al.(2018)Bhagwat, Viviano, Voineskos, and and]{Bhagwat18}
Bhagwat, N., Viviano, J.~D., Voineskos, A.~N., and and, M. M.~C.
\newblock Modeling and prediction of clinical symptom trajectories in
  alzheimer's disease using longitudinal data.
\newblock \emph{{PLOS} Computational Biology}, 14\penalty0 (9):\penalty0
  e1006376, sep 2018.
\newblock \doi{10.1371/journal.pcbi.1006376}.

\bibitem[Blei et~al.(2017)Blei, Kucukelbir, and McAuliffe]{Blei2017}
Blei, D.~M., Kucukelbir, A., and McAuliffe, J.~D.
\newblock Variational inference: A review for statisticians.
\newblock \emph{Journal of the American Statistical Association}, 112\penalty0
  (518):\penalty0 859--877, April 2017.
\newblock \doi{10.1080/01621459.2017.1285773}.
\newblock URL \url{https://doi.org/10.1080/01621459.2017.1285773}.

\bibitem[Brunton \& Kutz(2019)Brunton and Kutz]{Brunton2019}
Brunton, S.~L. and Kutz, J.~N.
\newblock \emph{Data-driven science and engineering}.
\newblock Cambridge University Press, Cambridge, England, February 2019.

\bibitem[Brunton et~al.(2016)Brunton, Proctor, and Kutz]{Brunton16}
Brunton, S.~L., Proctor, J.~L., and Kutz, J.~N.
\newblock Discovering governing equations from data by sparse identification of
  nonlinear dynamical systems.
\newblock \emph{Proceedings of the National Academy of Sciences}, 113\penalty0
  (15):\penalty0 3932--3937, mar 2016.
\newblock \doi{10.1073/pnas.1517384113}.

\bibitem[Buzs{\'{a}}ki(2006)]{Buzski2006}
Buzs{\'{a}}ki, G.
\newblock \emph{Rhythms of the Brain}.
\newblock Oxford University Press, October 2006.
\newblock \doi{10.1093/acprof:oso/9780195301069.001.0001}.
\newblock URL \url{https://doi.org/10.1093/acprof:oso/9780195301069.001.0001}.

\bibitem[Champion et~al.(2019)Champion, Lusch, Kutz, and Brunton]{Champion19}
Champion, K., Lusch, B., Kutz, J.~N., and Brunton, S.~L.
\newblock Data-driven discovery of coordinates and governing equations.
\newblock \emph{Proceedings of the National Academy of Sciences}, 116\penalty0
  (45):\penalty0 22445--22451, October 2019.
\newblock \doi{10.1073/pnas.1906995116}.
\newblock URL \url{https://doi.org/10.1073/pnas.1906995116}.

\bibitem[Chen et~al.(2018)Chen, Rubanova, Bettencourt, and Duvenaud]{chen2018}
Chen, R. T.~Q., Rubanova, Y., Bettencourt, J., and Duvenaud, D.~K.
\newblock Neural ordinary differential equations.
\newblock In \emph{Advances in Neural Information Processing Systems},
  volume~31. Curran Associates, Inc., 2018.
\newblock URL
  \url{https://proceedings.neurips.cc/paper/2018/file/69386f6bb1dfed68692a24c8686939b9-Paper.pdf}.

\bibitem[Chen et~al.(2021)Chen, Amos, and Nickel]{Chen2020_ODE}
Chen, R. T.~Q., Amos, B., and Nickel, M.
\newblock Learning neural event functions for ordinary differential equations.
\newblock In \emph{International Conference on Learning Representations}, 2021.
\newblock URL \url{https://openreview.net/forum?id=kW_zpEmMLdP}.

\bibitem[Chen et~al.(2017)Chen, Shojaie, and Witten]{Chen17}
Chen, S., Shojaie, A., and Witten, D.~M.
\newblock Network reconstruction from high-dimensional ordinary differential
  equations.
\newblock \emph{Journal of the American Statistical Association}, 112\penalty0
  (520):\penalty0 1697--1707, aug 2017.
\newblock \doi{10.1080/01621459.2016.1229197}.

\bibitem[Cybenko(1989)]{Cybenko89}
Cybenko, G.
\newblock Approximation by superpositions of a sigmoidal function.
\newblock \emph{Mathematics of Control, Signals, and Systems}, 2\penalty0
  (4):\penalty0 303--314, dec 1989.
\newblock \doi{10.1007/bf02551274}.

\bibitem[Dezfouli et~al.(2018)Dezfouli, Morris, Ramos, Dayan, and
  Balleine]{Dezfouli18}
Dezfouli, A., Morris, R., Ramos, F., Dayan, P., and Balleine, B.~W.
\newblock Integrated accounts of behavioral and neuroimaging data using
  flexible recurrent neural network models.
\newblock \emph{NIPS}, may 2018.
\newblock \doi{10.1101/328849}.

\bibitem[Drew et~al.(2020)Drew, Mateo, Turner, Yu, and Kleinfeld]{DREW2020782}
Drew, P.~J., Mateo, C., Turner, K.~L., Yu, X., and Kleinfeld, D.
\newblock Ultra-slow oscillations in {fMRI} and resting-state connectivity:
  Neuronal and vascular contributions and technical confounds.
\newblock \emph{Neuron}, 107\penalty0 (5):\penalty0 782--804, 2020.
\newblock ISSN 0896-6273.
\newblock \doi{https://doi.org/10.1016/j.neuron.2020.07.020}.
\newblock URL
  \url{https://www.sciencedirect.com/science/article/pii/S089662732030564X}.

\bibitem[Duncker et~al.(2019)Duncker, Bohner, Boussard, and Sahani]{Duncker19}
Duncker, L., Bohner, G., Boussard, J., and Sahani, M.
\newblock Learning interpretable continuous-time models of latent stochastic
  dynamical systems.
\newblock volume~97 of \emph{Proceedings of Machine Learning Research}, pp.\
  1726--1734. PMLR, 09--15 Jun 2019.
\newblock URL \url{http://proceedings.mlr.press/v97/duncker19a.html}.

\bibitem[Durstewitz(2003)]{Durstewitz2003}
Durstewitz, D.
\newblock Self-organizing neural integrator predicts interval times through
  climbing activity.
\newblock \emph{The Journal of Neuroscience}, 23\penalty0 (12):\penalty0
  5342--5353, June 2003.
\newblock \doi{10.1523/jneurosci.23-12-05342.2003}.
\newblock URL \url{https://doi.org/10.1523/jneurosci.23-12-05342.2003}.

\bibitem[Durstewitz(2017{\natexlab{a}})]{Durstewitz}
Durstewitz, D.
\newblock \emph{Advanced data analysis in {N}euroscience}.
\newblock Springer, 2017{\natexlab{a}}.
\newblock ISBN 978-3-319-59974-8.

\bibitem[Durstewitz(2017{\natexlab{b}})]{Durstewitz17}
Durstewitz, D.
\newblock A state space approach for piecewise-linear recurrent neural networks
  for identifying computational dynamics from neural measurements.
\newblock \emph{PLOS}, 13\penalty0 (6):\penalty0 e1005542, jun
  2017{\natexlab{b}}.
\newblock \doi{10.1371/journal.pcbi.1005542}.

\bibitem[Durstewitz et~al.(2000)Durstewitz, Seamans, and
  Sejnowski]{Durstewitz00b}
Durstewitz, D., Seamans, J.~K., and Sejnowski, T.~J.
\newblock Neurocomputational models of working memory.
\newblock \emph{Nature Neuroscience}, 3\penalty0 (S11):\penalty0 1184--1191,
  nov 2000.
\newblock \doi{10.1038/81460}.

\bibitem[Funahashi \& Nakamura(1993)Funahashi and Nakamura]{Funahashi93}
Funahashi, K. and Nakamura, Y.
\newblock Approximation of dynamical systems by continuous time recurrent
  neural networks.
\newblock \emph{Neural Networks}, 6\penalty0 (6):\penalty0 801--806, jan 1993.
\newblock \doi{10.1016/s0893-6080(05)80125-x}.

\bibitem[Gorbach et~al.(2017)Gorbach, Bauer, and Buhmann]{Gorbach17}
Gorbach, N.~S., Bauer, S., and Buhmann, J.~M.
\newblock Scalable variational inference for dynamical systems.
\newblock In \emph{Advances in Neural Information Processing Systems},
  volume~30. Curran Associates, Inc., 2017.
\newblock URL
  \url{https://proceedings.neurips.cc/paper/2017/file/e71e5cd119bbc5797164fb0cd7fd94a4-Paper.pdf}.

\bibitem[Halpern et~al.(2018)Halpern, Tubridy, Wang, Gasser, Popp, Davachi, and
  Gureckis]{Halpern18}
Halpern, D., Tubridy, S., Wang, H.~Y., Gasser, C., Popp, P. J.~O., Davachi, L.,
  and Gureckis, T.~M.
\newblock Knowledge tracing using the brain.
\newblock \emph{PsyArXiv}, apr 2018.
\newblock \doi{10.31234/osf.io/fmj48}.

\bibitem[Hanson \& Raginsky(2020)Hanson and Raginsky]{Hanson2020}
Hanson, J. and Raginsky, M.
\newblock Universal simulation of stable dynamical systems by recurrent neural
  nets.
\newblock In \emph{Proceedings of the 2nd Conference on Learning for Dynamics
  and Control}, volume 120 of \emph{Proceedings of Machine Learning Research},
  pp.\  384--392, The Cloud, 10--11 Jun 2020. PMLR.
\newblock URL \url{http://proceedings.mlr.press/v120/hanson20a.html}.

\bibitem[Haynes(2015)]{HAYNES2015}
Haynes, J.-D.
\newblock A primer on pattern-based approaches to fmri: Principles, pitfalls,
  and perspectives.
\newblock \emph{Neuron}, 87\penalty0 (2):\penalty0 257--270, 2015.
\newblock ISSN 0896-6273.
\newblock \doi{https://doi.org/10.1016/j.neuron.2015.05.025}.
\newblock URL
  \url{https://www.sciencedirect.com/science/article/pii/S0896627315004328}.

\bibitem[Haynes \& Rees(2006)Haynes and Rees]{Haynes2006}
Haynes, J.-D. and Rees, G.
\newblock Decoding mental states from brain activity in humans.
\newblock \emph{Nature Reviews Neuroscience}, 7\penalty0 (7):\penalty0
  523--534, July 2006.
\newblock \doi{10.1038/nrn1931}.
\newblock URL \url{https://doi.org/10.1038/nrn1931}.

\bibitem[Hern\'andez et~al.(2018)Hern\'andez, Wayment-Steele, Sultan, Husic,
  and Pande]{vaeComplexDynamics}
Hern\'andez, C.~X., Wayment-Steele, H.~K., Sultan, M.~M., Husic, B.~E., and
  Pande, V.~S.
\newblock Variational encoding of complex dynamics.
\newblock \emph{Phys. Rev. E}, 97:\penalty0 062412, Jun 2018.
\newblock \doi{10.1103/PhysRevE.97.062412}.
\newblock URL \url{https://link.aps.org/doi/10.1103/PhysRevE.97.062412}.

\bibitem[Hernandez et~al.(2018)Hernandez, Moretti, Wei, Saxena, Cunningham, and
  Paninski]{hernandez2018}
Hernandez, D., Moretti, A.~K., Wei, Z., Saxena, S., Cunningham, J., and
  Paninski, L.
\newblock Nonlinear evolution via spatially-dependent linear dynamics for
  electrophysiology and calcium data.
\newblock \emph{Neurons, Behavior, Data analysis, and Theory (NBDT)}, 2018.
\newblock URL \url{arXiv:1811.02459}.

\bibitem[Hert{\"a}g et~al.(2014)Hert{\"a}g, Durstewitz, and Brunel]{Hertaeg14}
Hert{\"a}g, L., Durstewitz, D., and Brunel, N.
\newblock Analytical approximations of the firing rate of an adaptive
  exponential integrate-and-fire neuron in the presence of synaptic noise.
\newblock \emph{Frontiers in Computational Neuroscience}, 8, sep 2014.
\newblock \doi{10.3389/fncom.2014.00116}.

\bibitem[Hochreiter et~al.(2007)Hochreiter, Heusel, and Obermayer]{btm247}
Hochreiter, S., Heusel, M., and Obermayer, K.
\newblock {Fast model-based protein homology detection without alignment}.
\newblock \emph{Bioinformatics}, 23\penalty0 (14):\penalty0 1728--1736, 05
  2007.
\newblock ISSN 1367-4803.
\newblock \doi{10.1093/bioinformatics/btm247}.
\newblock URL \url{https://doi.org/10.1093/bioinformatics/btm247}.

\bibitem[Iakovlev et~al.(2021)Iakovlev, Heinonen, and
  L{\"a}hdesm{\"a}ki]{iakovlev2021learning}
Iakovlev, V., Heinonen, M., and L{\"a}hdesm{\"a}ki, H.
\newblock Learning continuous-time {\{}pde{\}}s from sparse data with graph
  neural networks.
\newblock In \emph{International Conference on Learning Representations}, 2021.
\newblock URL \url{https://openreview.net/forum?id=aUX5Plaq7Oy}.

\bibitem[Kalman(1960)]{Kalman1960}
Kalman, R.~E.
\newblock A new approach to linear filtering and prediction problems.
\newblock \emph{Transactions of the ASME--Journal of Basic Engineering},
  82\penalty0 (Series D):\penalty0 35--45, 1960.

\bibitem[Kandala et~al.(2001)Kandala, Lang, Klasen, and
  Fahrmeir]{Fahrmeier2001}
Kandala, N.~B., Lang, S., Klasen, S., and Fahrmeir, L.
\newblock Semiparametric analysis of the socio-demographic and spatial
  determinants of undernutrition in two african countries, 2001.
\newblock URL
  \url{http://nbn-resolving.de/urn/resolver.pl?urn=nbn:de:bvb:19-epub-1626-2}.

\bibitem[Kimura \& Nakano(1998)Kimura and Nakano]{Kimura98}
Kimura, M. and Nakano, R.
\newblock Learning dynamical systems by recurrent neural networks from orbits.
\newblock \emph{Neural Networks}, 11\penalty0 (9):\penalty0 1589--1599, dec
  1998.
\newblock \doi{10.1016/s0893-6080(98)00098-7}.

\bibitem[Kingma \& Ba(2015)Kingma and Ba]{Adam}
Kingma, D.~P. and Ba, J.
\newblock Adam: {A} method for stochastic optimization.
\newblock In \emph{{ICLR} 2015, San Diego, CA, USA, May 7-9, 2015, Conference
  Track Proceedings}, 2015.
\newblock URL \url{http://arxiv.org/abs/1412.6980}.

\bibitem[Kingma \& Welling(2013)Kingma and Welling]{Kingma13}
Kingma, D.~P. and Welling, M.
\newblock {A}uto-{E}ncoding {V}ariational {B}ayes.
\newblock \emph{Arxiv}, 2013.

\bibitem[Koch \& Seveg(2003)Koch and Seveg]{Koch03}
Koch, C. and Seveg, I.
\newblock \emph{Methods in Neuronal Modeling (Computational Neuroscience
  Series): From Ions to Networks (Computational Neuroscience) Second Edition}.
\newblock A Bradford Book, 2003.
\newblock ISBN 9780262517133.

\bibitem[Koppe et~al.(2014)Koppe, Gruppe, Sammer, Gallhofer, Kirsch, and
  Lis]{Koppe2014}
Koppe, G., Gruppe, H., Sammer, G., Gallhofer, B., Kirsch, P., and Lis, S.
\newblock Temporal unpredictability of a stimulus sequence affects brain
  activation differently depending on cognitive task demands.
\newblock \emph{NeuroImage}, 101, 07 2014.
\newblock \doi{10.1016/j.neuroimage.2014.07.008}.

\bibitem[Koppe et~al.(2018)Koppe, Guloksuz, Reininghaus, and
  Durstewitz]{Koppe18}
Koppe, G., Guloksuz, S., Reininghaus, U., and Durstewitz, D.
\newblock Recurrent neural networks in mobile sampling and intervention.
\newblock \emph{Schizophrenia Bulletin}, 45\penalty0 (2):\penalty0 272--276,
  nov 2018.
\newblock \doi{10.1093/schbul/sby171}.

\bibitem[Koppe et~al.(2019)Koppe, Toutounji, Kirsch, Lis, and
  Durstewitz]{Koppe19}
Koppe, G., Toutounji, H., Kirsch, P., Lis, S., and Durstewitz, D.
\newblock Identifying nonlinear dynamical systems via generative recurrent
  neural networks with applications to {fMRI}.
\newblock \emph{{PLOS} Computational Biology}, 15\penalty0 (8):\penalty0
  e1007263, aug 2019.

\bibitem[Krishnan et~al.(2015)Krishnan, Shalit, and Sontag]{Krishnan15}
Krishnan, R.~G., Shalit, U., and Sontag, D.
\newblock Deep kalman filters.
\newblock \emph{arXiv}, 2015.

\bibitem[Kurle et~al.(2019)Kurle, G\"{u}nnemann, and der Smagt]{Kurle2019}
Kurle, R., G\"{u}nnemann, S., and der Smagt, P.~V.
\newblock Multi-source neural variational inference.
\newblock \emph{Proceedings of the {AAAI} Conference on Artificial
  Intelligence}, 33:\penalty0 4114--4121, July 2019.
\newblock \doi{10.1609/aaai.v33i01.33014114}.
\newblock URL \url{https://doi.org/10.1609/aaai.v33i01.33014114}.

\bibitem[Kusner et~al.(2017)Kusner, Paige, and
  Hern{\'a}ndez-Lobato]{hernandez2017}
Kusner, M.~J., Paige, B., and Hern{\'a}ndez-Lobato, J.~M.
\newblock Grammar variational autoencoder.
\newblock volume~70 of \emph{Proceedings of Machine Learning Research}, pp.\
  1945--1954. PMLR, 06--11 Aug 2017.
\newblock URL \url{http://proceedings.mlr.press/v70/kusner17a.html}.

\bibitem[Lahat et~al.(2015)Lahat, Adali, and Jutten]{Lahat15}
Lahat, D., Adali, T., and Jutten, C.
\newblock Multimodal data fusion: An overview of methods, challenges, and
  prospects.
\newblock \emph{Proceedings of the {IEEE}}, 103\penalty0 (9):\penalty0
  1449--1477, sep 2015.
\newblock \doi{10.1109/jproc.2015.2460697}.

\bibitem[Lambert(1992)]{Lambert1992}
Lambert, D.
\newblock Zero-inflated poisson regression, with an application to defects in
  manufacturing.
\newblock \emph{Technometrics}, 34:\penalty0 1--14, 02 1992.
\newblock \doi{10.1080/00401706.1992.10485228}.

\bibitem[Lee et~al.(2021)Lee, Jain, Park, and Yun]{lee2021crossattentional}
Lee, J.-T., Jain, M., Park, H., and Yun, S.
\newblock Cross-attentional audio-visual fusion for weakly-supervised action
  localization.
\newblock In \emph{International Conference on Learning Representations}, 2021.
\newblock URL \url{https://openreview.net/forum?id=hWr3e3r-oH5}.

\bibitem[Lewis et~al.(2016)Lewis, Setsompop, Rosen, and Polimeni]{Lewis2016}
Lewis, L.~D., Setsompop, K., Rosen, B.~R., and Polimeni, J.~R.
\newblock Fast {fMRI} can detect oscillatory neural activity in humans.
\newblock \emph{Proceedings of the National Academy of Sciences}, 113\penalty0
  (43):\penalty0 E6679--E6685, October 2016.
\newblock \doi{10.1073/pnas.1608117113}.
\newblock URL \url{https://doi.org/10.1073/pnas.1608117113}.

\bibitem[Liang et~al.(2015)Liang, Li, Chen, and Zeng]{Liang15}
Liang, M., Li, Z., Chen, T., and Zeng, J.
\newblock Integrative data analysis of multi-platform cancer data with a
  multimodal deep learning approach.
\newblock \emph{{IEEE}/{ACM} Transactions on Computational Biology and
  Bioinformatics}, 12\penalty0 (4):\penalty0 928--937, jul 2015.
\newblock \doi{10.1109/tcbb.2014.2377729}.

\bibitem[Lin \& Jegelka(2018)Lin and Jegelka]{Lin18}
Lin, H. and Jegelka, S.
\newblock {ResNet} with one-neuron hidden layers is a universal approximator.
\newblock In \emph{Advances in Neural Information Processing Systems},
  volume~31. Curran Associates, Inc., 2018.
\newblock URL
  \url{https://proceedings.neurips.cc/paper/2018/file/03bfc1d4783966c69cc6aef8247e0103-Paper.pdf}.

\bibitem[Lorenz(1963)]{Lorenz1963}
Lorenz, E.~N.
\newblock Deterministic nonperiodic flow.
\newblock \emph{Journal of the Atmospheric Sciences}, 20\penalty0 (2):\penalty0
  130--141, March 1963.
\newblock \doi{10.1175/1520-0469(1963)020<0130:dnf>2.0.co;2}.
\newblock URL
  \url{https://doi.org/10.1175/1520-0469(1963)020<0130:dnf>2.0.co;2}.

\bibitem[Lu et~al.(2017{\natexlab{a}})Lu, Pathak, Hunt, Girvan, Brockett, and
  Ott]{Lu17}
Lu, Z., Pathak, J., Hunt, B., Girvan, M., Brockett, R., and Ott, E.
\newblock Reservoir observers: Model-free inference of unmeasured variables in
  chaotic systems.
\newblock \emph{Chaos: An Interdisciplinary Journal of Nonlinear Science},
  27\penalty0 (4):\penalty0 041102, apr 2017{\natexlab{a}}.
\newblock \doi{10.1063/1.4979665}.

\bibitem[Lu et~al.(2017{\natexlab{b}})Lu, Pu, Wang, Hu, and Wang]{LuWang17}
Lu, Z., Pu, H., Wang, F., Hu, Z., and Wang, L.
\newblock The expressive power of neural networks: A view from the width.
\newblock \emph{NIPS}, 2017{\natexlab{b}}.

\bibitem[Machens et~al.(2005)Machens, Romo, and Brody]{Machens2005}
Machens, C.~K., Romo, R., and Brody, C.~D.
\newblock Flexible control of mutual inhibition: A neural model of two-interval
  discrimination.
\newblock \emph{Science}, 307\penalty0 (5712):\penalty0 1121--1124, February
  2005.
\newblock \doi{10.1126/science.1104171}.
\newblock URL \url{https://doi.org/10.1126/science.1104171}.

\bibitem[Maier et~al.(2017)Maier, Loos, and Hasenauer]{Maier2017}
Maier, C., Loos, C., and Hasenauer, J.
\newblock Robust parameter estimation for dynamical systems from
  outlier-corrupted data.
\newblock \emph{Bioinformatics}, pp.\  btw703, January 2017.
\newblock \doi{10.1093/bioinformatics/btw703}.
\newblock URL \url{https://doi.org/10.1093/bioinformatics/btw703}.

\bibitem[Meeds et~al.(2019)Meeds, Roeder, Grant, Phillips, and
  Dalchau]{Roeder19}
Meeds, T., Roeder, G., Grant, P., Phillips, A., and Dalchau, N.
\newblock Efficient amortised {B}ayesian inference for hierarchical and
  nonlinear dynamical systems.
\newblock volume~97 of \emph{Proceedings of Machine Learning Research}, pp.\
  4445--4455. PMLR, 09--15 Jun 2019.
\newblock URL \url{http://proceedings.mlr.press/v97/meeds19a.html}.

\bibitem[Miotto et~al.(2016)Miotto, Li, Kidd, and Dudley]{Miotto16}
Miotto, R., Li, L., Kidd, B.~A., and Dudley, J.~T.
\newblock Deep patient: An unsupervised representation to predict the future of
  patients from the electronic health records.
\newblock \emph{Scientific Reports}, 6\penalty0 (1), may 2016.
\newblock \doi{10.1038/srep26094}.

\bibitem[Molano-Mazon et~al.(2018)Molano-Mazon, Onken, Piasini, and
  Panzeri]{panzeri2018}
Molano-Mazon, M., Onken, A., Piasini, E., and Panzeri, S.
\newblock Synthesizing realistic neural population activity patterns using
  generative adversarial networks.
\newblock In \emph{International Conference on Learning Representations}, 2018.
\newblock URL \url{https://openreview.net/forum?id=r1VVsebAZ}.

\bibitem[Monfared \& Durstewitz(2020{\natexlab{a}})Monfared and
  Durstewitz]{Monfared2020-nody}
Monfared, Z. and Durstewitz, D.
\newblock Existence of n-cycles and border-collision bifurcations in
  piecewise-linear continuous maps with applications to recurrent neural
  networks.
\newblock \emph{Nonlinear Dynamics}, 101\penalty0 (2):\penalty0 1037--1052,
  July 2020{\natexlab{a}}.
\newblock \doi{10.1007/s11071-020-05841-x}.
\newblock URL \url{https://doi.org/10.1007/s11071-020-05841-x}.

\bibitem[Monfared \& Durstewitz(2020{\natexlab{b}})Monfared and
  Durstewitz]{monfared2020}
Monfared, Z. and Durstewitz, D.
\newblock Transformation of {R}e{LU}-based recurrent neural networks from
  discrete-time to continuous-time.
\newblock volume 119 of \emph{Proceedings of Machine Learning Research}, pp.\
  6999--7009. PMLR, 13--18 Jul 2020{\natexlab{b}}.
\newblock URL \url{http://proceedings.mlr.press/v119/monfared20a.html}.

\bibitem[Ngiam et~al.(2011)Ngiam, Khosla, Kim, Nam, Lee, and Ng]{Ngiam11}
Ngiam, J., Khosla, A., Kim, M., Nam, J., Lee, H., and Ng, A.
\newblock Multimodal deep learning.
\newblock In \emph{Proceedings of the 28th International Conference on Machine
  Learning (ICML-11)}, ICML '11, pp.\  689--696, New York, NY, USA, June 2011.
  ACM.
\newblock ISBN 978-1-4503-0619-5.

\bibitem[Owen et~al.(2005)Owen, McMillan, Laird, and Bullmore]{Owen2005}
Owen, A.~M., McMillan, K.~M., Laird, A.~R., and Bullmore, E.
\newblock N-back working memory paradigm: A meta-analysis of normative
  functional neuroimaging studies.
\newblock \emph{Human Brain Mapping}, 25\penalty0 (1):\penalty0 46--59, 2005.
\newblock \doi{10.1002/hbm.20131}.
\newblock URL \url{https://doi.org/10.1002/hbm.20131}.

\bibitem[Pandarinath et~al.(2018)Pandarinath, O'Shea, Collins, Jozefowicz,
  Stavisky, Kao, Trautmann, Kaufman, Ryu, Hochberg, Henderson, Shenoy, Abbott,
  and Sussillo]{Pandarinath2018}
Pandarinath, C., O'Shea, D.~J., Collins, J., Jozefowicz, R., Stavisky, S.~D.,
  Kao, J.~C., Trautmann, E.~M., Kaufman, M.~T., Ryu, S.~I., Hochberg, L.~R.,
  Henderson, J.~M., Shenoy, K.~V., Abbott, L.~F., and Sussillo, D.
\newblock Inferring single-trial neural population dynamics using sequential
  auto-encoders.
\newblock \emph{Nature Methods}, 15\penalty0 (10):\penalty0 805--815, September
  2018.
\newblock \doi{10.1038/s41592-018-0109-9}.
\newblock URL \url{https://doi.org/10.1038/s41592-018-0109-9}.

\bibitem[Paninski et~al.(2009)Paninski, Ahmadian, Ferreira, Koyama, Rad, Vidne,
  Vogelstein, and Wu]{Paninski09}
Paninski, L., Ahmadian, Y., Ferreira, D.~G., Koyama, S., Rad, K.~R., Vidne, M.,
  Vogelstein, J., and Wu, W.
\newblock A new look at state-space models for neural data.
\newblock \emph{Journal of Computational Neuroscience}, 29\penalty0
  (1-2):\penalty0 107--126, aug 2009.
\newblock \doi{10.1007/s10827-009-0179-x}.

\bibitem[Pathak et~al.(2018)Pathak, Hunt, Girvan, Lu, and Ott]{Pathak2018}
Pathak, J., Hunt, B., Girvan, M., Lu, Z., and Ott, E.
\newblock Model-free prediction of large spatiotemporally chaotic systems from
  data: A reservoir computing approach.
\newblock \emph{Physical Review Letters}, 120\penalty0 (2), January 2018.
\newblock \doi{10.1103/physrevlett.120.024102}.
\newblock URL \url{https://doi.org/10.1103/physrevlett.120.024102}.

\bibitem[Press et~al.(2007)Press, Teukolsky, Vetterling, and Flannery]{Press07}
Press, W.~H., Teukolsky, S.~A., Vetterling, W.~T., and Flannery, B.~P.
\newblock \emph{{N}umerical {R}ecipes}.
\newblock Cambridge University Pr., 2007.
\newblock ISBN 0521880688.

\bibitem[Purdon et~al.(2010)Purdon, Lamus, Hamalainen, and Brown]{Purdon10}
Purdon, P., Lamus, C., Hamalainen, M., and Brown, E.
\newblock A state space approach to multimodal integration of simultaneously
  recorded {EEG} and {fMRI}.
\newblock In \emph{2010 {IEEE} International Conference on Acoustics, Speech
  and Signal Processing}. {IEEE}, mar 2010.
\newblock \doi{10.1109/icassp.2010.5494906}.

\bibitem[Radu et~al.(2016)Radu, Lane, Bhattacharya, Mascolo, Marina, and
  Kawsar]{Radu16}
Radu, V., Lane, N.~D., Bhattacharya, S., Mascolo, C., Marina, M.~K., and
  Kawsar, F.
\newblock Towards multimodal deep learning for activity recognition on mobile
  devices.
\newblock In \emph{Proceedings of the 2016 {ACM} International Joint Conference
  on Pervasive and Ubiquitous Computing Adjunct - {UbiComp}16}. {ACM} Press,
  2016.
\newblock \doi{10.1145/2968219.2971461}.

\bibitem[Raissi(2018)]{Raissi18}
Raissi, M.
\newblock Deep hidden physics models: Deep learning of nonlinear partial
  differential equations.
\newblock \emph{Journal of Machine Learning Research}, 19\penalty0
  (25):\penalty0 1--24, 2018.
\newblock URL \url{http://jmlr.org/papers/v19/18-046.html}.

\bibitem[Rajkomar et~al.(2018)Rajkomar, Oren, Chen, Dai, Hajaj, Hardt, Liu,
  Liu, Marcus, Sun, Sundberg, Yee, Zhang, Zhang, Flores, Duggan, Irvine, Le,
  Litsch, Mossin, Tansuwan, Wang, Wexler, Wilson, Ludwig, Volchenboum, Chou,
  Pearson, Madabushi, Shah, Butte, Howell, Cui, Corrado, and Dean]{Rajkomar18}
Rajkomar, A., Oren, E., Chen, K., Dai, A.~M., Hajaj, N., Hardt, M., Liu, P.~J.,
  Liu, X., Marcus, J., Sun, M., Sundberg, P., Yee, H., Zhang, K., Zhang, Y.,
  Flores, G., Duggan, G.~E., Irvine, J., Le, Q., Litsch, K., Mossin, A.,
  Tansuwan, J., Wang, D., Wexler, J., Wilson, J., Ludwig, D., Volchenboum,
  S.~L., Chou, K., Pearson, M., Madabushi, S., Shah, N.~H., Butte, A.~J.,
  Howell, M.~D., Cui, C., Corrado, G.~S., and Dean, J.
\newblock Scalable and accurate deep learning with electronic health records.
\newblock \emph{npj Digital Medicine}, 1\penalty0 (1), may 2018.
\newblock \doi{10.1038/s41746-018-0029-1}.

\bibitem[Razaghi \& Paninski(2019)Razaghi and Paninski]{Razaghi19}
Razaghi, H.~S. and Paninski, L.
\newblock Filtering normalizing flows.
\newblock In \emph{4th workshop on Bayesian Deep Learning (NeurIPS 2019)},
  2019.

\bibitem[Rezende et~al.(2014)Rezende, Mohamed, and Wierstra]{Rezende14}
Rezende, D.~J., Mohamed, S., and Wierstra, D.
\newblock Stochastic backpropagation and approximate inference in deep
  generative models.
\newblock volume~32 of \emph{Proceedings of Machine Learning Research}, pp.\
  1278--1286, Bejing, China, 22--24 Jun 2014. PMLR.
\newblock URL \url{http://proceedings.mlr.press/v32/rezende14.html}.

\bibitem[Risken(1984)]{Risken84}
Risken, H.
\newblock \emph{The Fokker-Planck Equation}.
\newblock Springer Berlin Heidelberg, 1984.
\newblock \doi{10.1007/978-3-642-96807-5}.

\bibitem[Roweis \& Ghahramani(2002)Roweis and Ghahramani]{Roweis02}
Roweis, S. and Ghahramani, Z.
\newblock Learning nonlinear dynamical systems using the
  expectation-maximization algorithm.
\newblock In \emph{Kalman Filtering and Neural Networks}, pp.\  175--220. John
  Wiley {\&} Sons, Inc., 2002.
\newblock \doi{10.1002/0471221546.ch6}.

\bibitem[Rudy et~al.(2019)Rudy, Brunton, and Kutz]{Rudy19}
Rudy, S.~H., Brunton, S.~L., and Kutz, J.~N.
\newblock Smoothing and parameter estimation by soft-adherence to governing
  equations.
\newblock \emph{Journal of Computational Physics}, 398:\penalty0 108860, dec
  2019.
\newblock \doi{10.1016/j.jcp.2019.108860}.

\bibitem[Schmidt et~al.(2021)Schmidt, Koppe, Monfared, Beutelspacher, and
  Durstewitz]{SchmidtICLR}
Schmidt, D., Koppe, G., Monfared, Z., Beutelspacher, M., and Durstewitz, D.
\newblock Identifying nonlinear dynamical systems with multiple time scales and
  long-range dependencies.
\newblock In \emph{International Conference on Learning Representations}, 2021.
\newblock URL \url{https://openreview.net/forum?id=_XYzwxPIQu6}.

\bibitem[Shi et~al.(2019)Shi, N, Paige, and Torr]{Shi2019}
Shi, Y., N, S., Paige, B., and Torr, P.
\newblock Variational mixture-of-experts autoencoders for multi-modal deep
  generative models.
\newblock In \emph{Advances in Neural Information Processing Systems},
  volume~32. Curran Associates, Inc., 2019.
\newblock URL
  \url{https://proceedings.neurips.cc/paper/2019/file/0ae775a8cb3b499ad1fca944e6f5c836-Paper.pdf}.

\bibitem[Shi et~al.(2021)Shi, Paige, Torr, and N]{multimodalGenModels}
Shi, Y., Paige, B., Torr, P., and N, S.
\newblock Relating by contrasting: A data-efficient framework for multimodal
  generative models.
\newblock In \emph{International Conference on Learning Representations}, 2021.
\newblock URL \url{https://openreview.net/forum?id=vhKe9UFbrJo}.

\bibitem[Smith \& Brown(2003)Smith and Brown]{Brown03}
Smith, A. and Brown, E.
\newblock Estimating a {S}tate-{S}pace {M}odel from {P}oint {P}rocess
  {O}bservations.
\newblock \emph{Neural Computation}, 15, 2003.

\bibitem[Srivastava \& Salakhutdinov(2012)Srivastava and
  Salakhutdinov]{Srivastsava12}
Srivastava, N. and Salakhutdinov, R.~R.
\newblock Multimodal learning with deep boltzmann machines.
\newblock In \emph{Advances in Neural Information Processing Systems 25}, pp.\
  2222--2230. Curran Associates, Inc., 2012.

\bibitem[Strogatz(2018)]{Strogatz}
Strogatz, S.~H.
\newblock \emph{Nonlinear Dynamics and Chaos}.
\newblock {CRC} Press, may 2018.
\newblock \doi{10.1201/9780429492563}.

\bibitem[Sui et~al.(2012)Sui, Adali, Yu, Chen, and Calhoun]{Sui12}
Sui, J., Adali, T., Yu, Q., Chen, J., and Calhoun, V.~D.
\newblock A review of multivariate methods for multimodal fusion of brain
  imaging data.
\newblock \emph{Journal of Neuroscience Methods}, 204\penalty0 (1):\penalty0
  68--81, feb 2012.
\newblock \doi{10.1016/j.jneumeth.2011.10.031}.

\bibitem[Sushko \& Gardini(2010)Sushko and Gardini]{SUSHKO2010}
Sushko, I. and Gardini, L.
\newblock Degenerate bifurcations and border collisions in piecewise smooth
  {1D} and {2D} maps.
\newblock \emph{International Journal of Bifurcation and Chaos}, 20\penalty0
  (07):\penalty0 2045--2070, July 2010.
\newblock \doi{10.1142/s0218127410026927}.
\newblock URL \url{https://doi.org/10.1142/s0218127410026927}.

\bibitem[Sutter et~al.(2021)Sutter, Daunhawer, and Vogt]{multimodalElbo}
Sutter, T.~M., Daunhawer, I., and Vogt, J.~E.
\newblock Generalized multimodal {ELBO}.
\newblock In \emph{International Conference on Learning Representations}, 2021.
\newblock URL \url{https://openreview.net/forum?id=5Y21V0RDBV}.

\bibitem[Trischler \& D'Eleuterio(2016)Trischler and D'Eleuterio]{Trischler16}
Trischler, A.~P. and D'Eleuterio, G.~M.
\newblock Synthesis of recurrent neural networks for dynamical system
  simulation.
\newblock \emph{Neural Networks}, 80:\penalty0 67--78, aug 2016.
\newblock \doi{10.1016/j.neunet.2016.04.001}.

\bibitem[Tsai et~al.(2019)Tsai, Liang, Zadeh, Morency, and
  Salakhutdinov]{YaoHung2019}
Tsai, Y.~H., Liang, P.~P., Zadeh, A., Morency, L., and Salakhutdinov, R.
\newblock Learning factorized multimodal representations.
\newblock In \emph{7th International Conference on Learning Representations,
  {ICLR} 2019, New Orleans, LA, USA, May 6-9, 2019}. OpenReview.net, 2019.
\newblock URL \url{https://openreview.net/forum?id=rygqqsA9KX}.

\bibitem[Turner et~al.(2013)Turner, Forstmann, Wagenmakers, Brown, Sederberg,
  and Steyvers]{Turner13}
Turner, B.~M., Forstmann, B.~U., Wagenmakers, E.-J., Brown, S.~D., Sederberg,
  P.~B., and Steyvers, M.
\newblock A bayesian framework for simultaneously modeling neural and
  behavioral data.
\newblock \emph{{NeuroImage}}, 72:\penalty0 193--206, may 2013.
\newblock \doi{10.1016/j.neuroimage.2013.01.048}.

\bibitem[Vedantam et~al.(2018)Vedantam, Fischer, Huang, and
  Murphy]{vedantam2018generative}
Vedantam, R., Fischer, I., Huang, J., and Murphy, K.
\newblock Generative models of visually grounded imagination.
\newblock In \emph{International Conference on Learning Representations}, 2018.
\newblock URL \url{https://openreview.net/forum?id=HkCsm6lRb}.

\bibitem[Vlachas et~al.(2018)Vlachas, Byeon, Wan, Sapsis, and
  Koumoutsakos]{Vlachas2018}
Vlachas, P.~R., Byeon, W., Wan, Z.~Y., Sapsis, T.~P., and Koumoutsakos, P.
\newblock Data-driven forecasting of high-dimensional chaotic systems with long
  short-term memory networks.
\newblock \emph{Proceedings of the Royal Society A: Mathematical, Physical and
  Engineering Sciences}, 474\penalty0 (2213):\penalty0 20170844, May 2018.
\newblock \doi{10.1098/rspa.2017.0844}.
\newblock URL \url{https://doi.org/10.1098/rspa.2017.0844}.

\bibitem[Wilson(1999)]{Wilson1999}
Wilson, H.~R.
\newblock \emph{Spikes, Decisions, and Actions: The Dynamical Foundations of
  Neuroscience}.
\newblock Oxford University Press, 1 edition, 1999.

\bibitem[Woods(2010)]{Woods10}
Woods, A.~W.
\newblock Turbulent plumes in nature.
\newblock \emph{Annual Review of Fluid Mechanics}, 42\penalty0 (1):\penalty0
  391--412, jan 2010.
\newblock \doi{10.1146/annurev-fluid-121108-145430}.

\bibitem[Wu \& Goodman(2018)Wu and Goodman]{Wu2018MultimodalGM}
Wu, M. and Goodman, N.~D.
\newblock Multimodal generative models for scalable weakly-supervised learning.
\newblock In \emph{NeurIPS}, 2018.

\bibitem[Yu et~al.(2006)Yu, Afshar, Santhanam, Ryu, Shenoy, and Sahani]{Yu06}
Yu, B.~M., Afshar, A., Santhanam, G., Ryu, S.~I., Shenoy, K.~V., and Sahani, M.
\newblock Extracting dynamical structure embedded in neural activity.
\newblock In \emph{Advances in Neural Information Processing Systems 18}, pp.\
  1545--1552. MIT Press, 2006.

\bibitem[Zhao \& Park(2020)Zhao and Park]{Zhao2020}
Zhao, Y. and Park, I.~M.
\newblock Variational online learning of neural dynamics.
\newblock \emph{Frontiers in Computational Neuroscience}, 14, October 2020.
\newblock \doi{10.3389/fncom.2020.00071}.
\newblock URL \url{https://doi.org/10.3389/fncom.2020.00071}.

\end{thebibliography}
\bibliographystyle{icml2022}




\newpage

\appendix
\onecolumn
\section{Methodological Details}
\label{suppl:methodologicalDetails}

\subsection{EM algorithm: Optimization by Newton-Raphson Iterations} 
\label{suppl:Newton-Raphson}
Given the distributional assumptions of the observation and latent models, the joint log-likelihood in \eqref{GaussApp} spells out as:
\begin{align}
    \begin{split}
        \log \ &\text{p}_{\bm{\theta}}(\mathbf{Z},\mathbf{X},\mathbf{C})=\\
        &  - \frac{1}{2}(\mathbf{z}_1-\bm{\mu}_0 - \mathbf{F}\mathbf{s}_1)^\text{T}\bm{\Sigma}^{-1}(\mathbf{z}_1-\bm{\mu}_0 - \mathbf{F}\mathbf{s}_1)\\
        &- \frac{1}{2}\sum_{t=2}^T (\mathbf{z}_t-\mathbf{A}\mathbf{z}_{t-1}-\mathbf{W}\bm{\phi}(\mathbf{z}_{t-1})-\mathbf{h}-\mathbf{F}\mathbf{s}_t)^\text{T}\bm{\Sigma}^{-1}(\mathbf{z}_t-\mathbf{A}\mathbf{z}_{t-1}-\mathbf{W}\bm{\phi}(\mathbf{z}_{t-1})-\mathbf{h}-\mathbf{F}\mathbf{s}_t)\\
        &- \frac{1}{2}\sum_{t=1}^T\left(\mathbf{x}_t - \mathbf{B}\bm{\phi}(\mathbf{z}_{t})\right)^\text{T} \bm{\Gamma}^{-1}\left(\mathbf{x}_t - \mathbf{B}\bm{\phi}(\mathbf{z}_{t})\right)- \frac{T}{2}\left(\log |\bm{\Sigma}|+\log |\bm{\Gamma}|\right)\\
        &+\sum_{t=1}^{T} \left(\sum_{i=1}^{K-1} c_{it}\bm{\beta}_{i}\mathbf{z}_t - \log (1+\sum_{j=1}^{K-1}\exp(\bm{\beta}_j\mathbf{z}_t))\right)\\
        &+ \text{const.}
    \end{split}
    \label{JLL}
\end{align}
Let $\mathbf{z}=(z_{11},\dots,z_{M1},\dots,z_{1T},\dots,z_{MT})^T\in\mathbb{R}^{MT \times 1}$ be the concatenated vector of all state variables across all time steps, 
$\mathbf{d}_{\Omega}:=(d_{\Omega}^{(11)},d_{\Omega}^{(21)},\dots,d_{\Omega}^{(MT)})^T$ 
an indicator vector with $d_{\Omega}^{(mt)}=1 \ \forall z_{mt}>0$ and $d_{\Omega}^{(mt)}=0$ otherwise, and $\mathbf{D}_{\Omega}:= \text{diag}(\mathbf{d}_{\Omega})$. Arranging all terms quadratic, linear, and constant in $\mathbf{z}$ into big matrix form (cf. \cite{Koppe19}), 
one can rewrite optimization criterion (\ref{GaussApp}), (\ref{JLL}) as
\begin{align}
\begin{split}
Q^{\ast}_{\Omega}(\mathbf{Z})=&-\frac{1}{2}\left[2(\mathbf{a}^\text{T}\mathbf{a})- 2\bm{\Bar{\beta}}^\text{T}\mathbf{z} + \mathbf{z}^\text{T}\left(\mathbf{U}_0 + \mathbf{D}_{\Omega}\mathbf{U_1}+ \mathbf{U_1}^\text{T}\mathbf{D}_{\Omega} + \mathbf{D}_{\Omega} \mathbf{U_2}\mathbf{D}_{\Omega}\right)\mathbf{z}\right]\\
&~~~~~~~-\frac{1}{2}\left[-\mathbf{z}^\text{T}\left(\mathbf{v}_0 + \mathbf{D}_{\Omega}\mathbf{v}_1 \right) - \left(\mathbf{v}_0 + \mathbf{D}_{\Omega}\mathbf{v}_1 \right)^\text{T}\mathbf{z}\right]
\end{split}
\label{targetfunc}
\end{align}

where $\bm{\Bar{\beta}}=(\bm{\Bar{\beta}}_1,\dots,\bm{\Bar{\beta}}_T)\in\mathbb{R}^{MT\times 1}$ is a column vector with vector elements $\bm{\Bar{\beta}}_t:=[\bm{\beta}_1,\dots,\bm{\beta}_{K-1},\bm0]\bm{c}_t \in\mathbb{R}^{M\times 1}$ (picking out the regression vector $\bm{\beta}_l$ associated with the selected category $c_{lt}=1$ at time $t$), $\mathbf{a}=(\sqrt{\gamma_{\mathbf{z_{1}}}},\dots,\sqrt{\gamma_{\mathbf{z_{T}}}})^T\in\mathbb{R}^{T\times 1}$ with $\gamma_{\mathbf{z_{t}}}:=\log\left(1+\sum_{j=1}^{K-1}\exp(\bm{\beta}_j^T\mathbf{z}_t)\right)$. The components of the $\mathbf{U}$-matrices and $\mathbf{v}$-vectors are defined in \cite{Koppe19} and restated here (with slight model-specific adjustments) for convenience. All matrices $\mathbf{U}_{\{0,1,2\}}$ have a block-tridiagonal structure as in Eq. \ref{eq:Hessian} below with on- and off-diagonal blocks:
\begin{equation}
\quad 
\mathbf{U}_0^{tt} = \boldsymbol{\Sigma}^{-1}+ \mathbf{A}^\text{T}\boldsymbol{\Sigma}^{-1}\mathbf{A}, \quad
\mathbf{U}_0^{t+1,t} = -\boldsymbol{\Sigma}^{-1}\mathbf{A},\quad
\mathbf{U}_0^{t,t+1} = (\mathbf{U}_0^{t+1,t})^\text{T} \quad \text{for } t=1 \dots T-1,
\quad \mathbf{U}_0^{T,T} =  \boldsymbol{\Sigma}^{-1}
\end{equation}
\begin{equation}
\mathbf{U}_1^{t,t} = \mathbf{W}^\text{T}\boldsymbol{\Sigma}^{-1}\mathbf{A}, \quad 
\mathbf{U}_1^{t,t+1} = -\mathbf{W}^\text{T}\boldsymbol{\Sigma}^{-1} \quad \text{for } t=1 \dots T-1, \quad  \mathbf{U}_1^{T,T} = \mathbf{0}_{M\times M}
\end{equation}
\begin{equation}
\mathbf{U}_2^{t,t} = \mathbf{W}^\text{T}\boldsymbol{\Sigma}^{-1}\mathbf{W} + \mathbf{B}^\text{T}\boldsymbol{\Gamma}^{-1}\mathbf{B} \quad \text{for } t=1 \dots T-1, \quad  \mathbf{U}_2^{T,T} = \mathbf{B}^\text{T}\boldsymbol{\Gamma}^{-1}\mathbf{B}
\end{equation}

The vectors $\mathbf{v}_{\{0,1\}}$ are given by:
\begin{equation}
\mathbf{v}_0 = [
(\boldsymbol{\Sigma}^{-1}\mathbf{F}\mathbf{s}_1-\mathbf{A}^\text{T}\boldsymbol{\Sigma}^{-1}(\mathbf{F}\mathbf{s}_2+\mathbf{h}) + \boldsymbol{\Sigma}^{-1}\boldsymbol{\mu}_0)^\text{T},
\dots,
(\boldsymbol{\Sigma}^{-1}(\mathbf{F}\mathbf{s}_t+\mathbf{h})-\mathbf{A}^\text{T}\boldsymbol{\Sigma}^{-1}(\mathbf{F}\mathbf{s}_{t+1}+\mathbf{h}))^\text{T}, 
\dots,
(\boldsymbol{\Sigma}^{-1}(\mathbf{F}\mathbf{s}_T+\mathbf{h}))^\text{T}]^\text{T}
\end{equation}
\begin{equation}
\begin{split}
\mathbf{v}_1 = [&
(\mathbf{B}^\text{T} \boldsymbol{\Gamma}^{-1}\mathbf{x}_1 -\mathbf{W}^\text{T}\boldsymbol{\Sigma}^{-1}(\mathbf{F}\mathbf{s}_2+\mathbf{h}))^\text{T},
\dots,
(\mathbf{B}^\text{T} \boldsymbol{\Gamma}^{-1}\mathbf{x}_{t
}-\mathbf{W}^\text{T}\boldsymbol{\Sigma}^{-1}(\mathbf{F}\mathbf{s}_{t+1}+\mathbf{h}))^\text{T},
\dots,\\
&(\mathbf{B}^\text{T} \boldsymbol{\Gamma}^{-1}\mathbf{x}_{T-1}-\mathbf{W}^\text{T}\boldsymbol{\Sigma}^{-1}(\mathbf{F}\mathbf{s}_T+\mathbf{h}))^\text{T},(\mathbf{B}^\text{T} \boldsymbol{\Gamma}^{-1}\mathbf{x}_{T})^\text{T}]^\text{T}
\end{split}
\label{v_vec}
\end{equation}

In the original formulation of the PLRNN algorithm \cite{Durstewitz17,Koppe19}, the non-Gaussian terms due to $\mathrm{p}(\mathbf{C}\mid\mathbf{Z})$ were lacking, and hence criterion Eq.~(\ref{targetfunc}) was piecewise quadratic (owing to the piecewise linear ReLU activation) and could be addressed by an efficient fixed-point-iteration algorithm that alternates between (i) solving the set of equations $dQ^{\ast}_{\Omega}(\mathbf{Z})/d\mathbf{Z}=0$ linear in $\mathbf{Z}$ and (ii) recomputing the ReLU-derivatives $\mathbf{D}_{\Omega}$. Here we need to modify this algorithm, as the derivatives $dQ^{\ast}_{\Omega}(\mathbf{Z})/d\mathbf{Z}$ are not linear anymore, even for a fixed matrix $\mathbf{D}_{\Omega}$. Luckily, however, Eq.~(\ref{targetfunc}) will still be piecewise concave (within each orthant of the objective function landscape) and hence we can efficiently solve it using the Newton-Raphson (NR) scheme
\begin{align}
\mathbf{z}^{\mathrm{new}}=\mathbf{z}^{\mathrm{old}} - \alpha_{\mathrm{z}}~\mathbf{H}^{-1}\nabla \mathbf{z}^{\mathrm{old}}
\label{update}
\end{align}
where $\mathbf{D}_{\Omega}$ is recomputed after a few NR steps. Due to the Markov property of model Eq.~\ref{encoder}, the Hessian $\mathbf{H}:= \partial^2 Q^{\ast}_{\Omega}(\mathbf{Z})/\partial \mathbf{z}\partial \mathbf{z}^T$ has a specific, block-tridiagonal structure (see Eq. \eqref{eq:Hessian} and \citet{Koppe19}). This can be exploited (a) to store $\mathbf{H}$ in sparse format and (b) to obtain the inverse in $\mathcal{O}(T)$ time \cite{Paninski09}. The on- and off-diagonal blocks in $\mathbf{H}$ are composed of the components specified above (i.e., in Eqn. (\ref{targetfunc}) and below).

As noted in sect. \ref{sec:EM}, given the maximizer of Eq. (\ref{targetfunc}) as an estimate of the mean and the Hessian $\mathbf{H}$, we can then compute the required expectations $\mathrm{E}[\mathbf{z}]$, $\mathrm{E}[\mathbf{z}\mathbf{z}^T]$, $\mathrm{E}[\phi(\mathbf{z})]$, $\mathrm{E}[\mathbf{z}\phi(\mathbf{z})^T]$ and $\mathrm{E}[\phi(\mathbf{z})\phi(\mathbf{z})^T]$ mostly analytically \cite{Durstewitz17,Koppe19}, based on which parameters $\bm{\theta}_{\text{lat}}=\{\bm{\mu}_0,\mathbf{A},\mathbf{W},\mathbf{F},\mathbf{h},\bm{\Sigma}\}$ and $\bm{\theta}_X=\{\mathbf{B},\bm{\Sigma}\}$ can be solved for analytically as a linear regression problem. For the parameters $\bm{\beta}$ in Eq. (\ref{gtheta}) this is not possible, but since the problem is still concave efficient NR updates can again be used to solve numerically by
\begin{align}
    \bm{\beta}^{\text{new}}=\bm{\beta}^{\text{old}} - \alpha_{\beta} \bm{J}^{-1} \nabla \bm{\beta}^{\text{old}}~,
\label{fullNR}
\end{align}
where $\nabla \bm{\beta}^{\text{old}}:=\frac{\partial f(\bm{\theta})}{\partial \bm{\beta}}$ indicates the gradient, the Hessian is given by $\bm{J}$, and $\alpha_{\beta}$ is a learning rate (set to $\alpha_{\beta}=0.001$ here). Using the analytical derivations and approximations outlined here and in sect. \ref{sec:EM}, both the E- and the M-step become reasonably fast.

\subsection{Stepwise Training Protocol}\label{suppl:stepwiseTraining}

It has been shown previously \cite{Koppe19} that the approximation of the true underlying DS is strongly improved by embedding the EM algorithm (described in section \ref{sec:EM}) into a stepwise training protocol that successively shifts the burden of reproducing the observations from the observation model $\mathrm{p}_{\mathbf{\theta}}(\mathbf{X},\mathbf{C}\mid\mathbf{Z})$, as defined by Eqs.~(\ref{decoderA} \& \ref{decoderB}), to the latent process model $\mathrm{p}_{\mathbf{\theta}}(\mathbf{Z})$ as defined by Eq.~(\ref{encoder}). In a first step, a linear dynamical system (LDS) is trained by EM on the time series to find a suitable initial guess of parameters and states. 
Next, by deliberately fixing the covariance terms $\mathbf{\Gamma}$ and $\mathbf{\Sigma}$ of the observation and latent models, respectively, to certain values in the first full PLRNN runs, efficient training of the observation model is encouraged. In later steps, the observation model term $\mathrm{E}\left[\log \mathrm{p}_{\mathbf{\theta}}(\mathbf{X},\mathbf{C}\mid\mathbf{Z})\right]$ in optimization criterion Eq.~(\ref{ELBO}) is clamped off completely, thus enforcing the temporal consistency requirements in Eq.~(\ref{encoder}) and hence forcing the latent dynamical model to capture the observed temporal evolution in its own prior dynamics $\mathrm{p}_{\mathbf{\theta}}(\mathbf{Z})$. We use the same strategy here. For further details please refer to \cite{Koppe19}.

\subsection{Parameterization Approximate Posterior}
\label{suppl:paramterization}
Owing to the latent model's Markov property, the $MT \times MT$ Hessian $\mathbf{H}$ of the approximate posterior $\mathrm{q}(\mathbf{Z}\mid\mathbf{X},\mathbf{C})$ in Eq. (\ref{GaussApp}) and Eq. (\ref{eq:approximatePosterior}) has a specific block-tridiagonal structure (see also  \cite{Paninski09, Archer15}):

\begin{equation}
\mathbf{H} = 
\begin{pmatrix}
\mathbf{S}_1 & \mathbf{K}_1 & 0 & \cdots & \cdots & 0 \\
\mathbf{K}_1^\text{T} & \mathbf{S}_2 &  \mathbf{K}_2 & 0 & \cdots & 0 \\
0 & \mathbf{K}_2^\text{T} & \mathbf{S}_3 &  \mathbf{K}_3 & 0 & \vdots  \\
\vdots & 0 & \ddots & \ddots & \ddots & 0  \\
\vdots & \vdots & 0 & \mathbf{K}^\text{T}_{T-2} & \mathbf{S}_{T-1} & \mathbf{K}_{T-1}\\
0 & 0 & 0 & 0 & \mathbf{K}^\text{T}_{T-1} & \mathbf{S}_T
\end{pmatrix}~
\label{eq:Hessian}
\end{equation}
with $M \times M$ on-diagonal blocks $\mathbf{S}_t$ and off-diagonal blocks $\mathbf{K}_t$. 

For VI, we closely follow \cite{Archer15} who factorize the approximate posterior into a product of two Gaussians as
\begin{align}\label{eq:productOfGaussians}
        \mathrm{p}_{\boldsymbol\theta}(\mathbf{Z}\mid\mathbf{X},\mathbf{C}) &\approx q_{\boldsymbol\zeta} (\mathbf{Z}\mid\mathbf{X},\mathbf{C}) \propto \mathrm{q}_1(\mathbf{Z}\mid\mathbf{X},\mathbf{C})~\mathrm{q}_0(\mathbf{Z})~, \\
        q_{\boldsymbol\zeta} (\mathbf{Z}\mid\mathbf{X},\mathbf{C}) &= \mathcal{N}\left(\bm \mu_Z (\mathbf X,\mathbf{C})´, \mathbf \Lambda_Z (\mathbf X,\mathbf{C})\right)~,
\end{align}
with
\begin{align}\label{eq:q1}
    \mathrm{q}_1(\mathbf{Z}\mid\mathbf{X},\mathbf{C}) &= \mathcal{N}\left(\mathbf{m}_{{\boldsymbol\zeta}_\mu}, \mathbf{E}_{{\boldsymbol\zeta}_\Lambda} \right)~, \\
    \mathrm{q}_0(\mathbf{Z}) &= \mathcal{N}\left(0, \mathbf D \right)~.
\end{align}
Combining these two Gaussians yields for the combined mean $\bm{\mu}_Z$ and covariance $\bm\Lambda_Z$ of the posterior \cite{Archer15}
\begin{align}\label{eq:CovarianceParameterization}
    \mathbf \Lambda_Z (\mathbf X,\mathbf{C}) &= \left(\mathbf D^{-1} + \mathbf E^{-1}_{{{\boldsymbol\zeta}_\Lambda} }(\mathbf X,\mathbf{C}) \right)^{-1}~, \\
    \label{eq:CovarianceParameterization2}
    \bm \mu_Z (\mathbf X,\mathbf{C}) &= \mathbf \Lambda_Z(\mathbf X,\mathbf{C}) \mathbf E^{-1}_{{{\boldsymbol\zeta}_\Lambda}}(\mathbf X,\mathbf{C}) \mathbf{m}_{{\boldsymbol\zeta}_\mu}(\mathbf X,\mathbf{C})~,
\end{align}
where both $\mathbf D^{-1}$ and $\mathbf E^{-1}_{{{\boldsymbol\zeta}_\Lambda}}$ are $MT \times MT$ matrices of the general form given in Eq. \eqref{eq:Hessian}. Because of this block-tri-diagonal form, matrix inversions can be done efficiently in $\mathcal{O}(T)$ time \cite{Paninski09}. 
 
The formulation in Eqs. (\ref{eq:productOfGaussians}-\ref{eq:CovarianceParameterization2}) allows to insert a smoothness prior into the posterior via $\mathrm{q}_0(\mathbf{Z})$. More specifically, the on-diagonal blocks $\mathbf{P}_t$ and off-diagonal blocks $\mathbf{K}$ of prior matrix $\mathbf D^{-1}$ are time-independent and given by (t{Archer15}) 
\begin{align}\label{eq:blocks}
    &\mathbf{P}_1 = \mathbf{Q}_0^{-1} + \mathbf{V}^\text{T} \mathbf{Q}^{-1} \mathbf{V}~, \\
   &\mathbf{P}_t = \mathbf{Q}^{-1} + \mathbf{V}^\text{T} \mathbf{Q}^{-1} \mathbf{V}, ~~~ t = 2, \ldots, T-1~, \\
    &\mathbf{P}_T = \mathbf{Q}^{-1}~,  \\
    &\mathbf{K} = -\mathbf{V}^\text{T}\mathbf{Q}^{-1}~,
    \label{eq:blocksLast}
\end{align}
where $\boldsymbol V \in \mathbb{R}^{M \times M}$, $\boldsymbol Q_0 \in \mathbb{R}^{M \times M}$ and $\boldsymbol Q \in \mathbb{R}^{M \times M}$ are full parameter matrices optimized during training. This specific formulation follows the Kalman filter-smoother equations \cite{Kalman1960, Paninski09} where matrix $\mathbf D^{-1}$ collects the covariance terms that come from the process model, and the specific form of Eqs. (\ref{eq:blocks}-\ref{eq:blocksLast}) ensures they are arranged in the correct way within the full Hessian Eq. \eqref{eq:Hessian}. Matrix $\mathbf E^{-1}_{{{\boldsymbol\zeta}_\Lambda}}$ in turn captures the time-dependent terms from the observation model which appear only in the blocks on the diagonal. These on-diagonal blocks $\mathbf{L}_t$ as well as the time-dependent mean in Eq. \eqref{eq:q1} are parameterized through MLP-type neural network as
\begin{align}
    \mathbf{m}_{{\boldsymbol\zeta}_\mu}^{(t)}(\mathbf{x}_t,\mathbf{c}_t) &= \text{NN}_{{\boldsymbol\zeta}_\mu}(\mathbf{x}_t,\mathbf{c}_t)~, \\
    \mathbf{L}_t(\mathbf{x}_t,\mathbf{c}_t)&=\text{NN}_{{\boldsymbol\zeta}_{\Lambda}}(\mathbf{x}_t,\mathbf{c}_t), 
    \label{eq:nnParameterization}
\end{align}
with $\mathbf{m}_{{\boldsymbol\zeta}_\mu}^{(t)} \in \mathbb{R}^{M \times 1}$ and $\mathbf{L}_t \in \mathbb{R}^{M \times M}$. The diagonal blocks in matrix Eq. \eqref{eq:Hessian} are then given by $\mathbf{S}_t=\mathbf{P}_t+\mathbf{L}_t$. In our case, each NN has two separate input layers for the two data modalities, followed by two hidden layers of dimension $d_\text{h}= 25$ each. The input and hidden layers of the respective data modality are fully connected. The third layer combines the two input streams (therefore has input dimension $d_\text{c}=2 \cdot d_\text{h} = 50$), followed by an additional hidden layer, again fully connected. For all but the output layer we use ReLU activation functions. 

The VI code was implemented in Python/PyTorch, while for EM we modified a previous MatLab (MathWorks Inc.) implementation \cite{Koppe19}. All experiments were run on CPU-based servers (Intel Xeon Plat $8160$ @ $2.1$GHz with 24 cores or Intel Xeon Gold $6148$ @ $2.4$GHz with 20 cores), and the EM and VI algorithms were comparable in runtime (around $400$ minutes on a single core, i.e. without parallelization, for training on one Lorenz data set as used in sects. \ref{subsec:noisyLorenz} and \ref{sec:missDim}). 

\subsection{Hyper-parameter Selection}
\label{suppl:Hypers}
\paragraph{Expectation-Maximization}
In general, the EM algorithm does not have many hyper-parameters that are critical to tune. The most crucial one is the number of latent states $M$, which we checked for the Lorenz benchmarks within the range $M \in \{12,15,16,18\}$ based on previous work \citep{Koppe19}. $M = 15$ was selected as it led to the best state space reconstructions as assessed by our Kullback-Leibler measure $D_{KL}$ (see sect. \ref{suppl:kullback}) averaged across $20$ trajectories.  
Furthermore, a 
grid search was performed to determine the optimal Newton-Raphson learning rates in Eqn. \ref{update} and \ref{fullNR} across the ranges $\alpha_z=\{0.6,0.8,1\}$ (E-step) and $\alpha_{\beta}=\{0.0005,0.001,0.005,0.01\}$ (M-step), respectively. To strike a balance between good model fit, assessed by the model's ELBO approximation, 
and runtime (given the many samples drawn to create the cumulative histograms), values were chosen at which the ELBO did not notably increase anymore yet runtime was comparatively low. 

For the fMRI data, we again followed \citet{Koppe19} for orientation and searched for an optimal number of latent states within $M \in \{15,20,25,30\}$. $M =20$ was selected as it yielded the lowest $n=10$ step ahead prediction error (cf. Fig. \ref{ExpEval}A). 
Similar as for the Lorenz benchmarks, a grid search for the optimal Newton-Raphson learning rates was carried out within $\alpha_z \in \{0.6,0.8,1\}$ (E-step) and $\alpha_{\beta} \in \{0.0001, 0.00025, 0.0005,0.00075, 0.001\}$ (M-step; for $\alpha_{\beta}>0.001$ NR steps did not converge). As for the Lorenz benchmarks, parameters were selected that achieved a compromise between a reasonably high ELBO (without much further improvement for lower rates) and runtime (given the many runs across subjects and left-out sets that needed to be performed). 
In general, models were trained starting from five different initial conditions for the parameters, and from those retaining the configuration which yielded the highest likelihood. 
\paragraph{Variational Inference}
For comparability with the EM algorithm, the number of latent dimensions $M$ was fixed to be the same for VI. For the encoder network we explored a number of different settings with respect to performance and runtime: number of hidden layers $n_h \in \{2, 3, 4, 5\}$, number of units per layer $d_h \in \{10, 15, 25, 50, 75, 100\}$, learning rate $\alpha_v \in \{0.002, 0.001, 0.01\}$, as well as multi-layer perceptron (MLP) and convolutional neural network (CNN) architectures. Keeping runtime at a minimum and still obtaining good performance resulted in the parameter settings $n_h=2$, $d_h=25$, $\alpha_v=0.001$ and the MLP architecture. For the multimodal setup, an additional concatenation layer followed by another hidden layer was used. 
\paragraph{LSTM}
Hyper-parameters of the LSTM were chosen according to the lowest average $\mathrm{MSE}$ across $n$-step ahead predictions over the range $n \in \{5,7,9,11,13,15\}$. For comparability, a similar range for the number of LSTM cell states was explored as for the PLRNN's latent space dimensionality, 
$M = \{15,17,20,21,23,25\}$, for which $M=20$ turned out to yield the best performance. For all other hyper-parameters, the ranges explored followed \citet{btm247}. Specifically, optimal learning rates and the number of training epochs were determined by probing the ranges $\lambda \in \{0.001,0.002,0.003,0.004,0.005,0.006,0.007,0.008,0.009,0.01\}$ and $\#\mathrm{epochs} \in \{25,50,75,100,125,150,175,200\}$, respectively. 






\section{Details on Experimental Setups and Performance Measures}

\subsection{Lorenz Equations} 
\label{suppl:Lorenz}
The (stochastic) 3D-Lorenz system \cite{Lorenz1963} is defined by the set of equations

\begin{align}
    \begin{split}
        \mathrm{d}\mathrm{x}_1&=s(\mathrm{x}_2-\mathrm{x}_1)\mathrm{d}t+\mathrm{d}\epsilon_1(t)\\
        \mathrm{d}\mathrm{x}_2&=\left(\mathrm{x}_1(r-\mathrm{x}_3)-\mathrm{x}_2\right)\mathrm{d}t+\mathrm{d}\epsilon_2(t)\\
        \mathrm{d}\mathrm{x}_3&=\left(\mathrm{x}_1\mathrm{x}_2-b \mathrm{x}_3\right)\mathrm{d}t+\mathrm{d}\epsilon_3(t) ~.
    \end{split}
    \label{Lorenz}
\end{align}
The system was solved with $4$\textit{th} order Runge-Kutta numerical integration. Process noise was injected by adding an \textit{i.i.d.} Gaussian term $\mathrm{d}\boldsymbol{\epsilon} \sim \mathcal{N}\left(0, 0.0025 \times \mathrm{d}t \mathbf{I}\right)$ to the three equations. Parameter values used here were $s=10$, $r=28$, and $b=8/3$, which place the Lorenz system into the chaotic regime.

\subsection{Agreement in Attractor Geometries}
\label{suppl:kullback}
\begin{figure*}
    \begin{center}
    \includegraphics[width=0.9\textwidth]{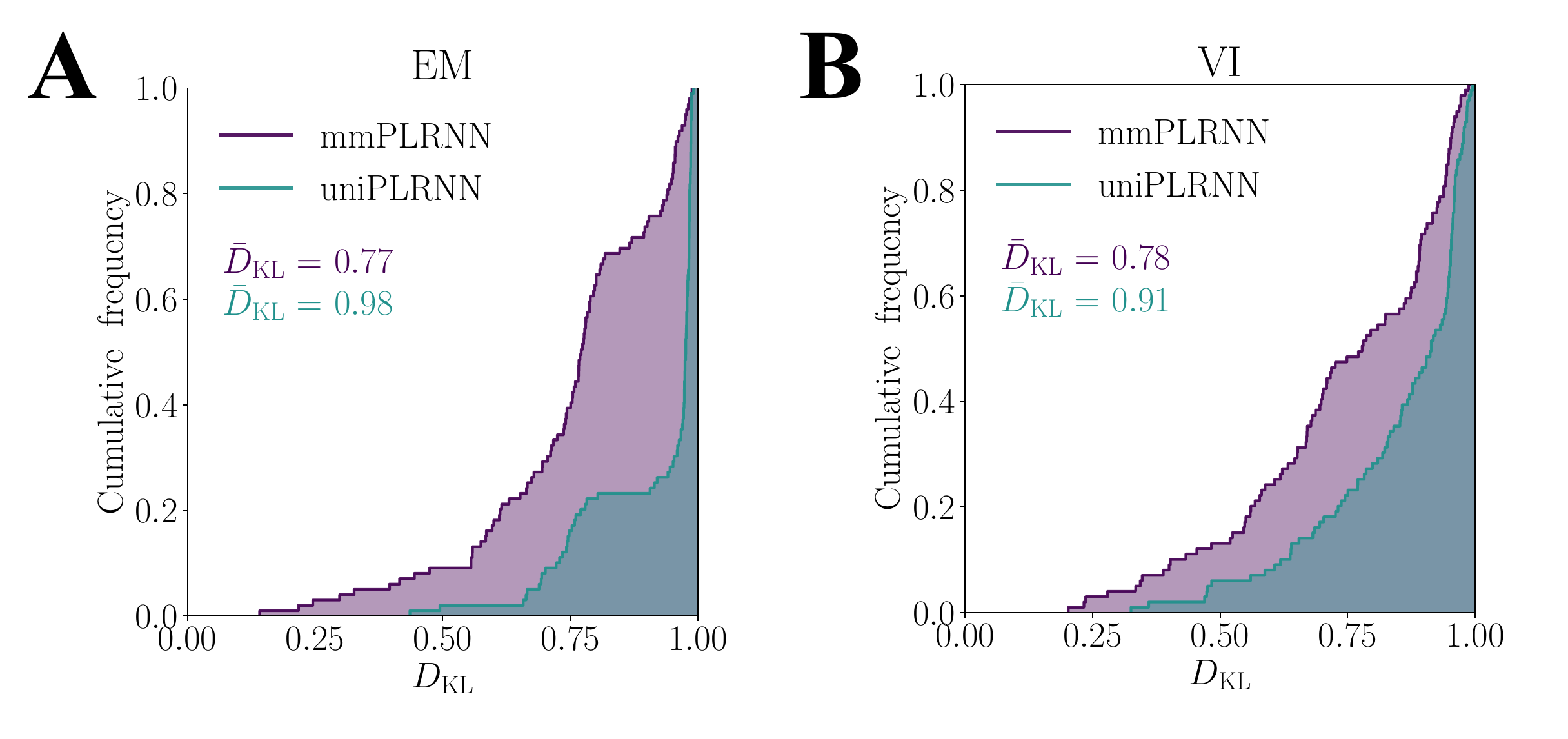}
    \end{center}
\caption{
    Same setup as in Fig. \ref{NoisyLorenz}, but with one Lorenz variable ($x_2$) omitted from the observations. A) Cumulative performance histograms ($n=100$ runs) as in Fig. \ref{NoisyLorenz}C for continuous+categorical (purple) vs. only (degenerated) continuous (cyan) information channel for models trained by EM.  $\Bar{D}_\text{KL}$ indicates the median. B) Same as A for models trained by VI. }
\label{fig:ReducedLorenz}
\end{figure*}

Following \citet{Koppe19} and \citet{SchmidtICLR}, we quantify the agreement in attractor geometries by comparing the true and model-generated probability distributions across observations $\mathbf{X}$ in state space through a Kullback-Leibler divergence ($D_{\text{KL}}$), approximated as
\begin{align}
D_{\text{KL}}(\hat{\mathrm{p}}_{\text{true}}^{(k)}(\mathbf{x})\mid\mid\hat{\mathrm{p}}_{\text{gen}}^{(k)}(\mathbf{x}\mid\mathbf{z}))\approx\sum_{k=1}^K \hat{\mathrm{p}}_{\text{true}}^{(k)}(\mathbf{x}) \log\left(\frac{\hat{\mathrm{p}}_{\text{true}}^{(k)}(\mathbf{x})}{\hat{\mathrm{p}}_{\text{gen}}^{(k)}(\mathbf{x}\mid\mathbf{z})}\right)
\label{KLdis}
\end{align}
where $\hat{\mathrm{p}}_{\text{true}}(\mathbf{x})$ is the true distribution of observations across state space, $\hat{\mathrm{p}}_{\text{gen}}(\mathbf{x}\mid\mathbf{z})$ the simulated distribution generated by the (freely running) PLRNN, and index $k$ runs across bins in state space. See \citet{Koppe19} or \citet{SchmidtICLR} for more details. To evaluate $D_{\text{KL}}$, trajectories of length $T= 100~000$ were generated from both ground truth system and PLRNN, from which the initial transients ($500$ time points) were cut off. To yield a relative measure in $[0,1]$, $D_{\text{KL}}$ was normalized to the largest $D_{\text{KL}}^{max} \approx 18.4$ across all iterations from both the noisy and incomplete Lorenz experiments using either EM or VI (i.e., same maximum value was used for all graphs in Figs. \ref{NoisyLorenz}, \ref{fig:ReducedLorenz}). 

In order to compute $D_{\text{KL}}$ in observation space for the case where one Lorenz variable was missing (see below), 
a projection from the latent into the \textit{full} observation space was computed by linear regression (i.e., re-computing matrix $\mathbf{B}$ from observation model Eq. (\ref{decoderA}) for the full set of observations).

\begin{figure}
    \centering
    \includegraphics[width=0.45\textwidth]{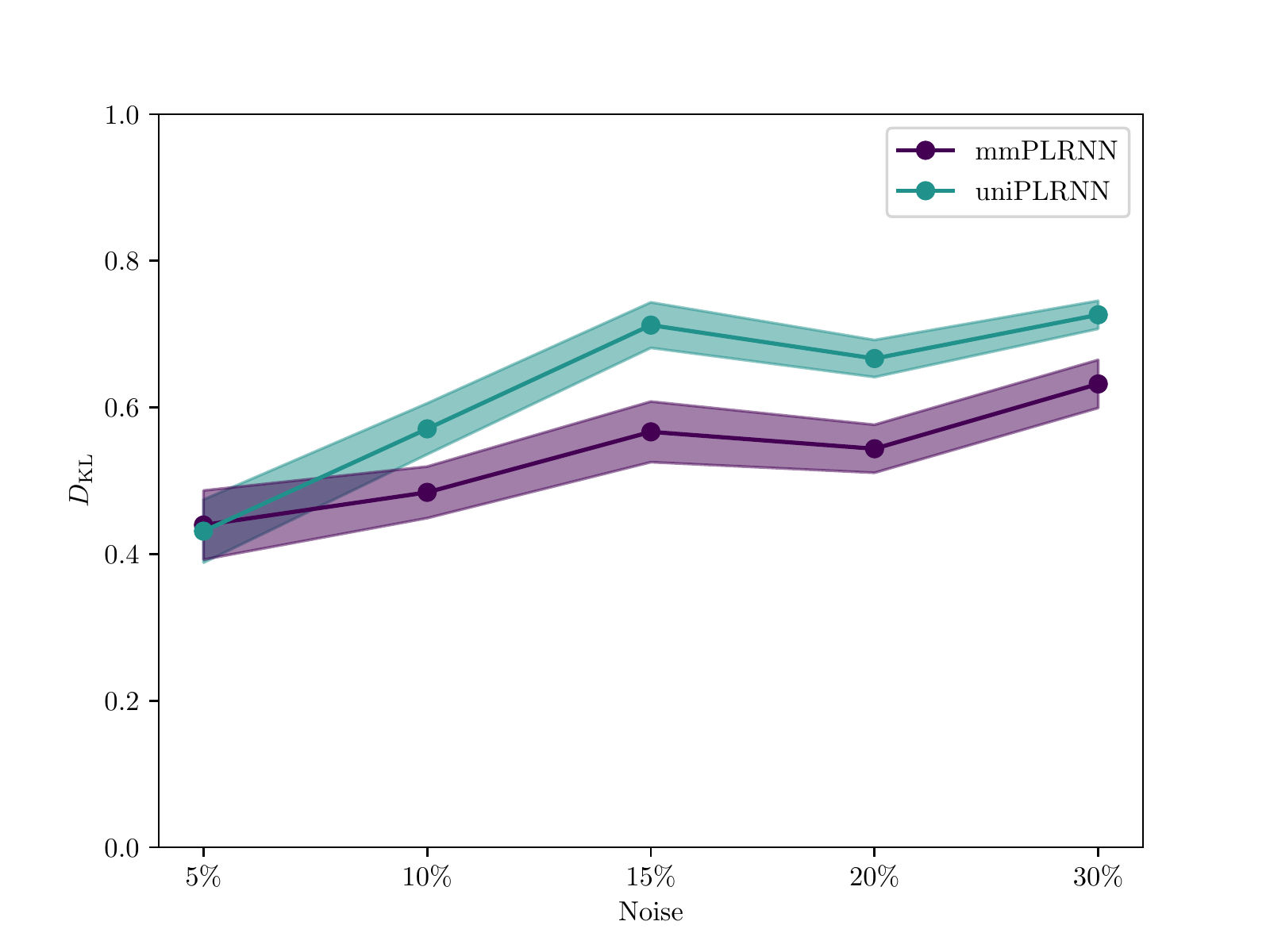}
    \caption{Mean $D_\text{KL}$ and SEM ($n=40$) for the uni- vs. mmPLRNN trained by VI on the noisy Lorenz benchmark across longer time series of $T=2500$ for different observation noise levels. Note that a) performance for VI at $10\%$ improved to levels comparable to EM for these longer time series as compared to the $T=1000$ series used in the main text (cf. Fig. \ref{NoisyLorenz}), and b) access to additional categorical information significantly improves reconstruction performance only as the continuous-Gaussian channel becomes more severely corrupted by noise, but not for low noise levels ($5\%$), as one may intuitively expect.}
    \label{fig:noiseLevels}
\end{figure}


\subsection{Details on fMRI Experiments and Analysis}
\label{suppl:fmriDetails}
Briefly, human study participants \cite{Koppe2014} were presented with a sequence of images of different shapes (rectangles and triangles) under three different task conditions while lying in a fMRI scanner: The continuous delayed response 1-back task (CDRT), the continuous matching 1-back task (CMT), and a 0-back control choice reaction task (CRT). In all three task stages subjects had to correctly indicate the type of stimulus currently presented (0-back) or 1 step before in the sequence (1-back). Task blocks were presented sequentially and repeated 5 times (amounting to $3\times5$ task blocks), and only differed w.r.t. the instruction phase and displayed response options. In addition to these three task phases, a resting condition where the participants were just lying still in the scanner with eyes closed, and an instruction phase which informed the participants about the upcoming task phase, were included in the experiments. These constituted the five task stages, each of which involving different mental processes, which were assigned different categorical labels for decoding. Any external information concerning the type of stimulus presented was omitted during training, in order to not provide the algorithm with any other source of information about the labels or dynamics. Analysis of BOLD signals was performed on the first principal components of 10 brain regions bilaterally relevant to the n-back task \cite{Owen2005} (yielding $N=20$ time series per subject). The details of the experimental setup are given in \cite{Koppe2014}; fully anonymized data were obtained from the authors of that study and used here with their permission.

The confusion matrix reported in Fig. \ref{ExpEval}C was determined through 5-fold cross-validation. Specifically, this was done by fixing the mmPLRNN’s parameters from the training set and obtaining posterior state estimates $\mathrm{E}[\mathbf{Z}^\mathrm{test}\mid\mathbf{X}^\mathrm{test}]$ from the left-out BOLD signal $\mathbf{X}^\mathrm{test}$ alone (using the pre-trained encoder model), after re-estimating the initial condition $\boldsymbol\mu_0$ on the left-out segment. These inferred latent trajectories were then used to predict the unseen task labels through the previously trained observation model $\mathrm{p}_{\mathbf{\theta}_\mathrm{cat}}(\mathbf{C}^\mathrm{test}\mid\mathbf{Z}^\mathrm{test})$  (Eq.~(\ref{decoderB})). 
Only validation blocks from all subjects were included for which the BOLD dynamics was reconstructed successfully on the respective training set (only in these cases the training was considered successful; note that the quality of the recordings may differ considerably among subjects). This yielded a total of $k=84$ left-out test sets from $l=21$ subjects. Relative frequencies were then computed by first summing across all these test sets from those subjects 
and then dividing by the respective total counts.

For all analyses in Figs. \ref{ExpEval} \& \ref{Fig4},
data from $l=21$ subjects were used, excluding $5$ subjects from a total of $26$ due to apparent artefacts in the the BOLD signal or model divergence (including subjects with artefacts did, however, not change the results of the analyses, i.e., significant non-/differences remained as reported in the graph). 

For the analysis of class label prediction (Fig.~\ref{ExpEval}C), all classifiers received the same $20$ BOLD principal components as inputs and the $5$ task class labels as to-be-predicted outputs as used for mmPLRNN training. For SVM, radial basis function (RBF) kernels with tradeoff parameter $C=1.0$ were used. All classifiers were trained using the \texttt{scikit-learn} library.

\subsection{MSE $n$-step ahead Prediction}

We define the MSE for $n$-step ahead predictions as
\begin{equation}
    \mathrm{MSE}(n) = \frac{1}{N \cdot (T-n)} \sum_{t=1}^{T-n} \left\|\mathbf{x}_{t+n} - \hat{\mathbf{x}}_{t+n} \right\|^2_2
\end{equation}
where $\hat{\mathbf{x}}_{t+n} \in \mathbb{R}^{N\times 1}$ is produced by iterating model Eq. (\ref{encoder}) $n$ steps forward in time from its current best estimator $\text{E}[\mathbf{z}_t|\mathbf{x}_{1:T}]$, and generating from this forward-iterated value $\hat{\mathbf{z}}_{t+n}$ the prediction $\hat{\mathbf{x}}_{t+n}:=\text{E}[\mathbf{x}_{t+n}|\hat{\mathbf{z}}_{t+n}]$ according to observation model Eq. (\ref{decoderA}). 

\subsection{Calculation of Lyapunov Exponents}
\label{sec:LyapExp}
Let us denote the PLRNN mean function as given in Eq. \ref{encoder} by $F_{\boldsymbol\theta}(\mathbf{z}_{t-1},\mathbf{s}_{t})$. Then the system's Jacobian at time $t$ is given by
\begin{align}\label{eq:jacobian}
\mathbf{J}_t \, := \, \frac{\partial F_{\boldsymbol\theta}(\mathbf{z}_{t-1},\mathbf{s}_{t})}{\partial \mathbf{z}_{t-1}} 
 =\, \mathbf{A} + \mathbf{W} \mathbf{D}_{\Omega(t-1)},
\end{align}
where $\mathbf{D}_{\Omega(t-1)}$ is a diagonal $\{0,1\}$ matrix with the $ReLU$ derivatives at time point $t-1$ on its diagonal, i.e. an $M\times M$ submatrix of $\mathbf{D}_{\Omega}$ as defined in Appx. \ref{suppl:Newton-Raphson}. The maximal Lyapunov exponent along a PLRNN trajectory $\{\mathbf{z}_1,\mathbf{z}_2,\cdots,\mathbf{z}_T,\cdots\}$ is then defined as
\begin{align}\label{eq:lyap}
\lambda_{max} \,:= \, 
\lim_{T \rightarrow \infty} \frac{1}{T}
\log   \left\| \prod_{r=0}^{T-2}  \mathbf{J}_{T-r} \right\|,
\end{align}
where ${\rVert \cdot \lVert}$ denotes the spectral norm (or any subordinate norm) of a matrix. Practically, one can let the orbit evolve until $\lambda_{max}$ in Eq. \ref{eq:lyap} converges for some $T$.

\subsection{Benchmarks: Lorenz System with Incomplete Gaussian and Categorical Observations}
\label{sec:missDim}
As a second test case, we studied whether additional categorical information (as in sect. 4.1, Fig.~\ref{NoisyLorenz}) could also help to identify the chaotic Lorenz system when one of its dynamical variables ($x_2$) was missing from the observations, i.e. only $\mathbf{x}^{\mathrm{red}}_t=(x_{1t},x_{3t})^T$ was provided. This is a nontrivial case, since the Lorenz system is a highly condensed minimal model for the chaotic attractor dynamics, i.e. with each variable absolutely necessary to produce the observed behavior (unlike many experimental systems which often have quite some redundancy, as in the nervous system or molecular networks). Yet, as shown in Fig.~\ref{fig:ReducedLorenz}, non-quantitative, categorical data could efficiently compensate for the lack of continuous time series information about one of the system’s variables. In terms of summary statistics, this is reflected in the $D_\text{KL}$ distributions (Fig.~\ref{fig:ReducedLorenz}) when the mmPLRNN inference was run with (purple) vs. without (cyan) access to categorical data on top of the linearly transformed $(x_{1t},x_{3t})$ time series.  
Again this was generally true for both EM and VI, with EM performing somewhat better on average.

\section{Examples of GLM-type Observation Models: Categorical, Gamma and Zero-Inflated Poisson Distributions}
\label{suppl:carlosModels}

For the categorical model, Eq. \ref{decoderB}, that we explored in the main text, the natural link function is given by
\begin{align}
    \begin{split}
        &\pi_i=\frac{\exp(\bm{\beta}_i^T\mathbf{z}_t)}{1+\sum_{j=1}^{K-1}\exp(\bm{\beta}_j^T\mathbf{z}_t)} \in [0,1]\quad \forall i\in\{1 \ldots K-1\}\\
        &\pi_K=\frac{1}{1+\sum_{j=1}^{K-1} \exp(\bm{\beta}_j^T\mathbf{z}_t)}~ \text{, such that}\quad \sum_{i=1}^{K}\pi_i=1,
    \end{split}
    \label{catprob}
\end{align}
where $\bm{\beta}_i\in \mathbb{R}^{M\times1}$ is the vector of regression weights for category $i={1 \dots K-1}$.

Here we illustrate the VI-based mmPLRNN on two further examples of observation models, namely when we have observations that could best be accounted for by 1) a gamma-distribution or by 2) a Poisson distribution with an excess of zeros (Zero-Inflated Poisson (ZIP) model \cite{Lambert1992}). Examples of the latter are 
event counts for earthquakes or a neuron's action potentials where occasional periods of increased activity may be separated by relatively long periods of silence.

Real-valued gamma time series $\mathbf{G} =\{ \mathbf{g}_t \}$, $ \mathbf{g}_t  \in\mathbb{R}_+^{J\times 1}$, $t=1\dots T$, would be described by the conditional density 
\begin{equation}
\mathbf{g}_t \mid \mathbf{z}_t \sim Gamma(\omega, \boldsymbol{\nu}_t)
\label{eq: gamma_model}
\end{equation}
where $\omega > 0$ is a shape parameter and $\boldsymbol{\nu}_t \in \mathbb{R}_+^{J\times 1}$ are scale parameters. We may connect them to the latent states $\mathbf{z}_t$ in model Eq.~(\ref{encoder}) through a GLM, where we model the distribution's conditional means $\boldsymbol{\mu}_t = (\mu_{1t}, \dots , \mu_{Jt})^T = \omega/\boldsymbol{\nu}_t$ at time $t$ via the log link function:
\begin{equation}
\log\mu_{jt} = \boldsymbol {\xi}_j^{T} \mathbf{z}_t \qquad \forall j\in\{1 \dots J\}
\label{eq: gamma_log_link}
\end{equation}
where $ \boldsymbol {\xi}_j \in \mathbb{R}^{M\times 1}$ is a vector of regression weights for each of the gamma observations $j=1 \dots J$. Note that like for the Gaussian and categorical models we did not include the usual constant offset (bias) term in the GLM, as we assume the overall level is determined by the latent states $\mathbf{z}_t$ which are equipped with their own bias terms $\mathbf{h}$ (cf. Eq.~(\ref{encoder}), avoiding model redundancy).

As an example of a somewhat more complex, composite distributional model, assume we have integer-valued Poisson data $\mathbf{Q} =\{ \mathbf{q}_t \}$, $ \mathbf{q}_t  \in\mathbb{N}^{L\times 1}$, $t=1\dots T$, but with an excess proportion of zeros. This situation could be described by the ZIP model \cite{Lambert1992} which assumes that each observation $q_{lt}$ at time $t$ is either 0 with probability $\psi_{lt}$, or distributed according to a Poisson process with mean $\lambda_{lt}$ with probability $1-\psi_{lt}$:
\begin{equation}
\mathbf{q}_t \mid \mathbf{z}_t \sim \mbox{ZIP}(\boldsymbol{\psi}_t, \boldsymbol{\lambda}_t) 
\label{eq: zip_model}
\end{equation}
\begin{equation}
\qquad
\mathrm{p}(q_{lt} \mid \mathbf{z}_t) =
\left\{
  \begin{array}{lr}
    \psi_{lt} + (1-\psi_{lt})e^{-\lambda_{lt}} & \qquad \mbox{for } q_{lt} = 0 \\\
    (1-\psi_{lt}) \frac{\lambda_{lt}^{q_{lt}}}{q_{lt}! } e^{-\lambda_{lt}} & \qquad \mbox{for } q_{lt} > 0
  \end{array}
\right.\\
\label{eq: zip_pdf}
\end{equation}
The elements of probability vector $\boldsymbol{\psi}_t = (\psi_{1t}, \dots , \psi_{Lt})^T$ are produced from the latent states  $\mathbf{z}_t$ through the logit link function
\begin{equation}
\log\frac{\psi_{lt}}{1-\psi_{lt}} =\boldsymbol {\eta}_l^{T} \mathbf{z}_t \qquad \forall l\in\{1 \dots L\},
\label{eq: zip_log_link_psi}
\end{equation}
where $ \boldsymbol {\eta}_l \in \mathbb{R}^{M\times 1}$ is a vector of regression weights for each Poissonian variable $l=1 \dots L$. Likewise, the mean values $\boldsymbol{\lambda}_t = (\lambda_{1t}, \dots , \lambda_{Lt})^\text{T}$ of the Poisson distribution are connected to the latent states $\mathbf{z}_t$ through the log-link function
\begin{equation}
\log\lambda_{lt} =\boldsymbol {\gamma}_l^{T} \mathbf{z}_t \qquad \forall l\in\{1 \dots L\},
\end{equation}
where $ \boldsymbol {\gamma}_l \in \mathbb{R}^{M\times 1}$ is another vector of regression weights for each of the Poisson observations $l=1 \dots L$.

Assuming, as for the Gaussian and categorical observations, that the gamma ($\mathbf{g}_t$) and ZIP ($\mathbf{q}_t $) observations are conditionally independent given the latent state $\mathbf{z}_t$, the log-likelihood for this setup including Gaussian data is given by the following factorization:
\begin{equation}
\begin{aligned}
\mathrm{p}_{\mathbf\theta} (\mathbf{x}, \mathbf{g}, \mathbf{q})  =& \int_{\mathbf{Z}}\mathrm{p}_{\mathbf{\theta}}(\mathbf{z}_1) \prod_{t=2}^{T}\mathrm{p}_{\mathbf{\theta}}(\mathbf{z}_t\mid\mathbf{z}_{t-1}) \prod_{t=1}^{T}\mathrm{p}_{\mathbf{\theta}}(\mathbf{x}_t\mid\mathbf{z}_t)\\
&
\prod_{t=1}^{T}\mathrm{p}_{\mathbf{\theta}}(\mathbf{q}_t\mid\mathbf{z}_t)\prod_{t=1}^{T} \mathrm{p}_{\mathbf{\theta}}(\mathbf{g}_t\mid\mathbf{z}_t)\mathrm{d}\mathbf{Z}
\end{aligned}
\label{eq:multimodal_loglik}
\end{equation}
with parameters $\boldsymbol \theta = \{\boldsymbol \mu_0, \mathbf{A}, \mathbf{W} , \mathbf{F}, \mathbf{h}, \mathbf \Sigma, \mathbf{B}, \mathbf \Gamma, \omega, \{\boldsymbol \xi_1,\dots,\boldsymbol \xi_J\},\allowbreak \{ \boldsymbol \eta_1,\dots,\boldsymbol\eta_L\}, \{\boldsymbol \gamma_1,\dots,\boldsymbol\gamma_L\} \}$.
Again we approximate this by the ELBO as given for categorical+Gaussian observations in Eq.~(\ref{ELBO}), and parameterize the variational approximation $\mathrm{q}_{\mathbf{\zeta}}(\mathbf{Z}\mid\mathbf{X},\mathbf{G}, \mathbf{Q})$ through neural networks in the very same way as described in Appx. \ref{suppl:paramterization} above.

\begin{figure*}[htb]
    \begin{center}
    \includegraphics[width=1.0\textwidth]{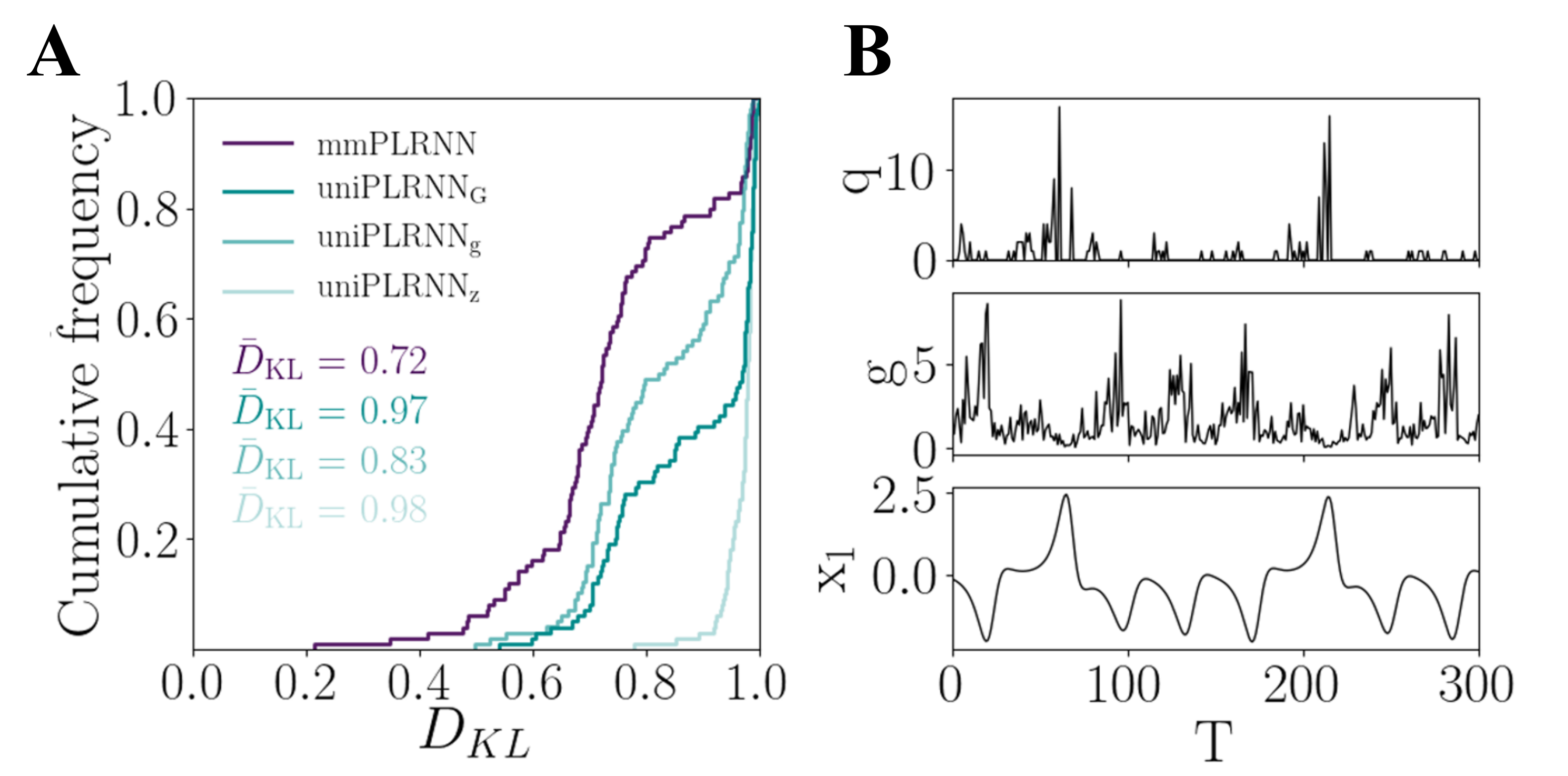}
    \end{center}
    \caption{
    A) Cumulative histograms ($n=100$ runs) of $D_\text{KL}/D_\text{KL}^{max}$ for all three (Gaussian, gamma, ZIP) uni-modal PLRNNs (from light to dark cyan) and the mmPLRNN connected to all three observation modalities simultaneously (purple). $\bar{D}_{KL}$ indicates the median. G = Gaussian, g = gamma, Z = ZIP model. 
    B) Example time series of ZIP (top) and gamma (center) observations generated from the Lorenz attractor dynamics for which the time series of one variable is shown at the bottom. Note that the original Lorenz time series information appears highly degraded in the sparse ZIP output, and the structure is also distorted in the gamma output.}
    \label{LorenzGammaPoisson}
\end{figure*}

Fig. \ref{LorenzGammaPoisson} illustrates the application to the noisy Lorenz setting (as described in sect. \ref{subsec:noisyLorenz}), this time with gamma and/or ZIP observations on top, or instead, of Gaussian observations. Simulations were run with $M=15$ latent states, a linear-Gaussian observation model in addition to the two models described above, and $20\%$ observation noise in the Gaussian channel. For the mmPLRNN including all 3 modalities, the NN parameterizing the approximate posterior had three separate input layers for the three data modalities, followed by two hidden layers of dimension $d_\text{h}= 25$ each. A third layer of dimension $d_\text{c}=75 = 3d_\text{h}$, where $3$ is the number of modalities, combined the modality-specific streams, followed by two additional hidden layers of size $d_\text{h}$. All layers were fully connected with ReLU activation, except for the output layer. As the graphs indicate, additional observations from other modalities again help to reconstruct the attractor geometry if the Gaussian observations are very noisy. We also observed, however, that it is difficult to reconstruct the Lorenz system from (by definition non-negative) gamma or ZIP observations alone. At least for ZIP-distributed data this is easy to explain (see example time series in Fig. \ref{LorenzGammaPoisson}B), as these - by model definition - low and sparse counts provide only very little information about the Lorenz attractor's complex geometry. 



\begin{figure*}[htb]
    \centering
    \includegraphics[width=\textwidth]{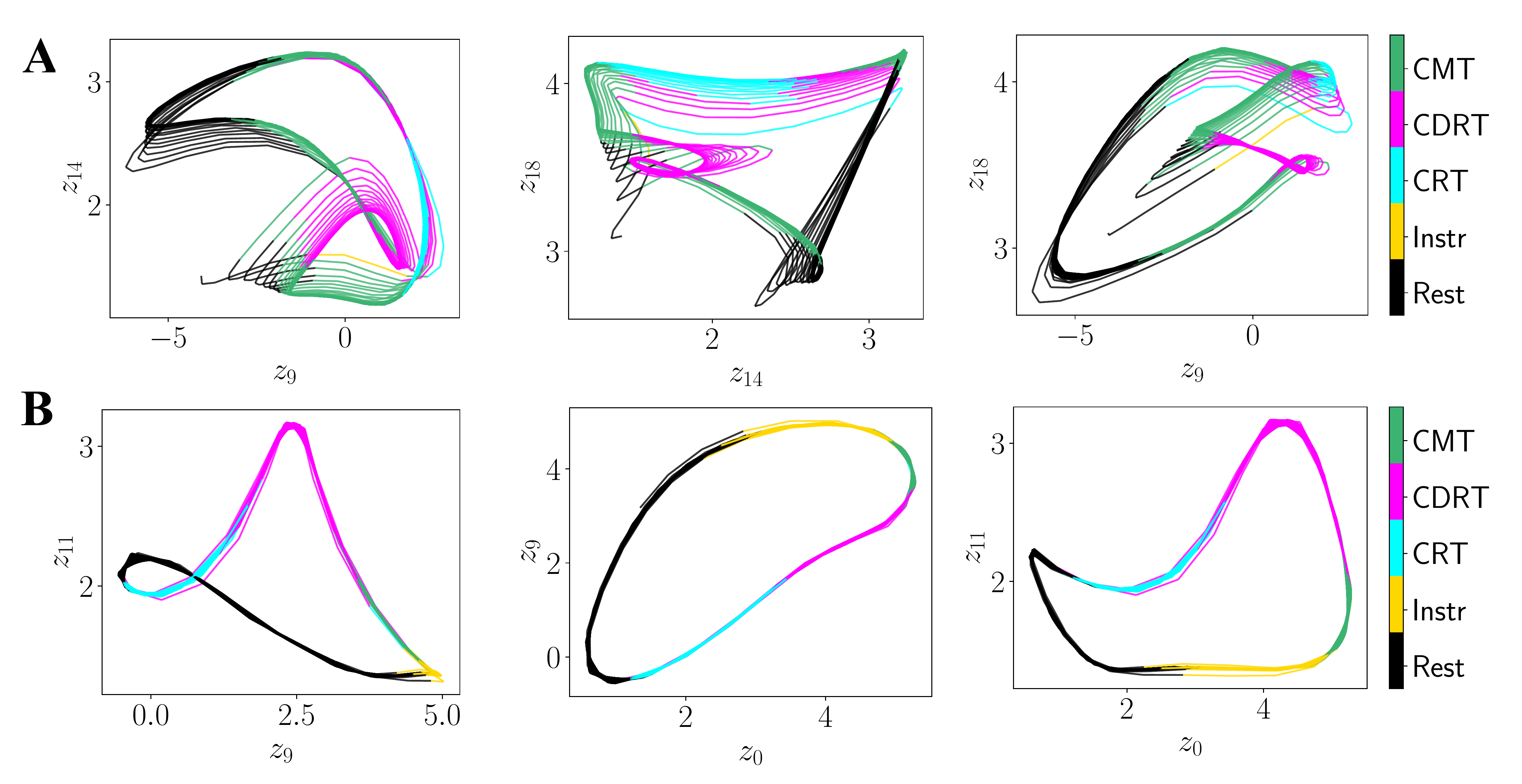}
    \caption{Two further examples for the association between predicted class labels (color coding) and learned BOLD dynamics. Shown are 2d subspaces of a mmPLRNN's generated state space. Subspaces chosen for display were selected according to Fisher's discriminant criterion.}
    \label{fig:stateSpaceExampleAppx1}
\end{figure*}

\end{document}